\documentclass[11pt,a4paper]{article}
\usepackage{times,latexsym, amssymb}
\usepackage{url}
\usepackage{multirow}
\usepackage[T1]{fontenc}
\usepackage{authblk}
\usepackage{pifont}

\usepackage[acceptedWithA]{tacl}
\usepackage{graphicx, multirow, makecell, booktabs, amsmath, arydshln}

\usepackage{xspace,mfirstuc,tabulary}

\newcommand{\name}{\textsc{DEIm}\xspace}
\newcommand{\SDM}{\textit{Edisa}\xspace}

\newif\iftaclinstructions
\taclinstructionsfalse 
\iftaclinstructions
\renewcommand{\confidential}{}
\renewcommand{\anonsubtext}{(No author info supplied here, for consistency with
TACL-submission anonymization requirements)}
\newcommand{\instr}
\fi

\title{Discover, Explain, Improve: An Automatic Slice Detection Benchmark for Natural Language Processing}

\author[1]{\textbf{Wenyue Hua}}
\author[2]{\textbf{Lifeng Jin}}
\author[2]{\textbf{Linfeng Song}}
\author[2]{\textbf{Haitao Mi}}
\author[1]{\\\textbf{Yongfeng Zhang}}
\author[2]{\textbf{Dong Yu}}
\affil{Rutgers University, New Brunswick\quad \textsuperscript{2}Tencent America}
\affil[1]{\ttfamily{{{wenyue.hua, yongfeng.zhang}@rutgers.edu}}}
\affil[2]{\ttfamily{{lifengjin, lfsong, haitaomi, dyu}@tencent.com}}

\begin{document}
\maketitle

\begin{abstract}
Pretrained natural language processing (NLP) models
have achieved high overall performance, but they still make systematic errors.
Instead of manual error analysis, research on slice detection models (SDM), which automatically identify underperforming groups of datapoints, has caught escalated attention in Computer Vision
for both understanding model behaviors and providing insights for future model training and designing.
However, little research on SDM and quantitative evaluation of their effectiveness have been conducted on NLP tasks.
Our paper fills the gap by proposing a benchmark named ``\textbf{D}iscover, \textbf{E}xplain, \textbf{Im}prove (\name)'' for classification NLP tasks along with a new SDM \SDM. \SDM discovers coherent and underperforming groups of datapoints; \name then unites them under human-understandable concepts and provides comprehensive evaluation tasks and corresponding quantitative metrics.
The evaluation in \name shows that \SDM can accurately select error-prone datapoints with informative semantic features that summarize error patterns. Detecting difficult datapoints directly boosts model performance without tuning any original model parameters, showing that discovered slices are actionable for users\footnote{Code and Benchmark are available here: \url{https://github.com/Wenyueh/DEIM}.}.
\end{abstract}

\section{Introduction}
While deep learning models \cite[inter alia]{BERT, roBERTa, electra} achieve high overall performance on many tasks, they often display systematic errors \cite{google, chatbot, recruit} correlated with biases, challenging data points, and data collection issues. Investigating these errors and their associated features is crucial for understanding models' strengths and weaknesses. Although manual error analysis is typically employed for identifying biases and erroneous behaviors, its efficiency and quality are limited. Consequently, automatic slice detection models (SDM) are motivated to streamline the analysis process by identifying systematic errors in any trained machine learning model \cite{domino, beyondaccuracy, modelagnostic, polyjuice}, based on the observation that representations of error instances may share features and thus similar to each other.

In SDM, a slice refers to a set of datapoints sharing a specific attribute. An error slice is a slice characterized by low accuracy \cite{domino}. Identifying these error slices serves three primary purposes: (1) locating error-prone datapoints to enable direct prediction adjustments, (2) gaining insights into model behavior to foster better comprehension and interpretation, and (3) guiding additional model training through strategies such as slice-specific modeling, data augmentation, and active learning. Therefore, an effective slice detection model should (1) accurately locate error-prone data points, (2) offer coherent error slices which help yield intelligible error-correlated features, and (3) enhance model performance when complemented with suitable tools.

\begin{figure*}
    \centering
    \includegraphics[scale=0.4]{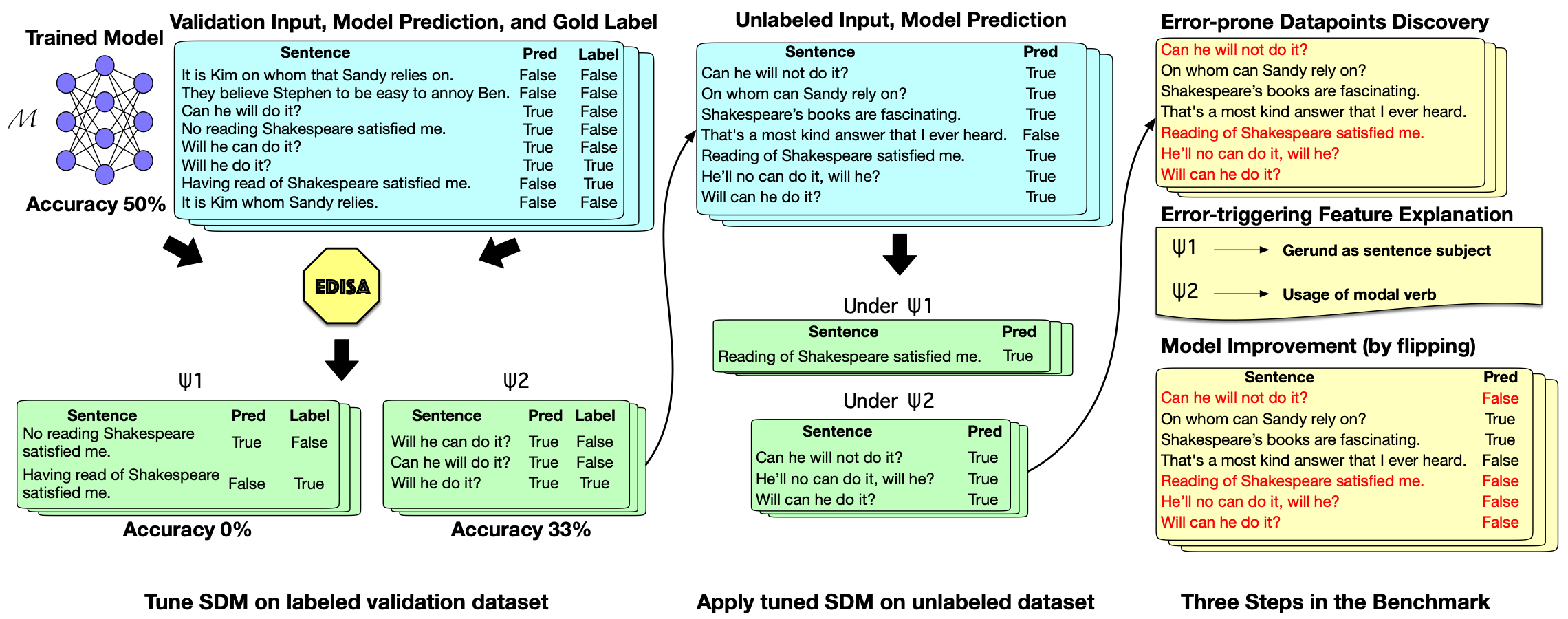}
    \caption{An SDM (\SDM) takes a trained model $\mathcal{M}$ and a labeled dataset as inputs. It is tuned on inputs, predictions, and labels. The tuned SDM generates multiple low-performing slicing functions such as $\psi_1$ and $\psi_2$. Applying the SDM on any unlabeled dataset will group the datapoints based on slicing functions. The Discover module collects error-prone datapoints; The Explain module assigns features to each error slice, and the Improve module enhances model performance. Here, it simply flips the predictions of identified error-prone datapoints.}
    \label{fig:DEI}
    \vspace{-10pt}
\end{figure*}

In this study, we introduce a comprehensive benchmark that assesses SDMs with three modules: namely Discover, Explain, and Improve, each of which corresponds to a point previously discussed. We also propose a new SDM \SDM to serve as a baseline on the benchmark. The usage of \SDM and the evaluation pipeline of the benchmark \name are depicted in Figure \ref{fig:DEI}. Here we briefly introduce the three modules:
\\
\textbf{Discover}: This module utilizes a tuned SDM to detect error-prone datapoints on any unlabeled datasets for a specific trained NLP model denoted as $\mathcal{M}$. The evaluation of the discovery capability is straightforward, which verifies whether the located error-prone datapoints are indeed mispredicted by $\mathcal{M}$. In the example in Figure \ref{fig:DEI}, one sentence classified to $\psi_1$ and three sentences classified to $\psi_2$ are deemed as error-prone. \\
\textbf{Explain}: This module employs linguistic tools to articulate why a model fails on a given error slice, consolidating the reasons into human-comprehensible concepts. For each identified error slice, it discerns linguistic features that occur substantially more often within the slice. These features potentially elucidate why the model inaccurately predicts these data points.
In Figure \ref{fig:DEI}, sentences in $\psi_1$ all contain gerund (verbal ending in -ing that functions as a noun) as sentence subject, indicating that it is likely to be the reason why these datapoints are mispredicted. In order to assess the cohesiveness of the discovered error slices, we evaluate measures such as homogeneity and completeness of each slice with respect to their error-correlated features. \\
\textbf{Improve}: This module showcases how model improvement is realized based on discovered error slices utilizing three techniques: selective prediction \cite{selective, towards}, flipping, and active learning. For instance, as shown in Figure \ref{fig:DEI}, we invert the prediction for each identified error-prone data point. These three model improvement methods also serve as external evaluations of SDMs. To verify the usefulness of the discovered error slices, we examine whether the model's performance escalates after implementing these techniques.

In the three-module benchmark \name, each module concentrates on one specific application of an SDM: (1) detection of error-prone data points, (2) interpretability of error slices, and (3) improvement of model performance. Each module provides evaluation tasks with the necessary tools and quantitative metrics. Each module incorporates evaluation tasks equipped with essential tools and quantitative metrics. Experimental results on \SDM indicate that it can effectively identify error-prone data points in unlabeled datasets and precisely detect error-correlated features, which contribute directly to enhanced model performance.

The paper is organized as follows: Section 2 discusses recent work on slice detection models; Section 3 introduces the model structure of \SDM model. Section 4 presents the details of the \name benchmark and all relevant tools. Section 5 presents experiment results and relevant ablation studies. Section 6 concludes this paper.

\section{Related Work}

Explainable model predictions are crucial in various research areas. Discovering error-correlated features in datapoints both increases model performance and delivers insights into future model design. In CV, research has been reported to use learned input representations to identify semantically meaningful slices where errors are made in prediction \cite{domino, spotlight, completeness, subclass, multiaccuracy, featureextraction}. \citet{domino} recently proposed the SOTA automatic error detection method DOMINO. 
In NLP, task-specific automatic error analysis research has been conducted on tasks such as document-level information extraction \cite{info-ex}, coreference resolution \cite{cofres}, and machine translation \cite{machinetrans}. There is also extensive research conducted on different model evaluations to see whether models make erroneous datapoints in certain types of noising datapoints \cite{synthetic, modelwild} or adversarial datapoints \cite{adversarialribeiro, adversarialparaphrase}. Another line of work
including \citet{cartography, Curriculum, ILDAE} focuses on evaluating the model-independent difficulty level of datapoints. Recently, \citet{seal} introduced an interactive visualization tool for underperforming slices using token-level features. 

However, as far as we know, there has not been a comprehensive evaluation benchmark that circumvents all the aspects of SDM in NLP. Therefore in this project, we contribute to the research area by designing a benchmark \name for all classification tasks: it provides (1) task-independent comprehensive linguistic feature benchmark for potential explanations, (2) quantitative experiments for both error slice quality and error-prone datapoints detection efficacy in unlabeled datasets, (3) and corresponding metrics that facilitate future development. We also propose a new SDM model \SDM which performs fairly well to serve as the SDM baseline for \name benchmark in NLP field. Its simple structure and promising results show a good prospect of this field.

\section{\SDM Model}

\SDM model is a new model that we proposed for slice detection in NLP. This section describes the model structure, training objective, and inference procedure of \SDM. Subsequently, we compare this model with the current state-of-the-at SDM model DOMINO \cite{domino} to underscore why such model structure design is necessary. 

In \SDM, we posit the existence of a set of $k$ interpretable slices, each distinguished by one or more crucial features that differentiate the slice from other data points. 
\SDM specifically focuses on error-correlated features, that is, features co-occurring with incorrect predictions. Thus, for the same task and dataset, the set of features and the $k$ slices may vary with respect to different NLP models. The objective of an SDM is to identify these $k$ slices for a trained $\mathcal{M}$ in an unsupervised manner. Ideally, the discovery of these $k$ slices requires a sufficiently large dataset where both input information and model prediction information are accessible. We mimic this setting by providing a labeled validation dataset, aiming to identify the $k$ slices within it.

To formally introduce the model, \SDM can be seen as a function $g$ that takes in a trained NLP model $\mathcal{M}$ and a labeled dataset $\mathcal{D}$ to generate $k$ slicing functions $\{\psi_i\}_{i=1}^{i=k}$:
\begin{multline}
    g(\mathcal{M}, \mathcal{D}) = \{\psi_i: D\times\mathcal{M} \to \{0,1\}\mid 1\leq i\leq k\}
\end{multline}

\subsection{\SDM's Model Structure}
\SDM is an \textbf{E}rror-\textbf{dis}tance-\textbf{a}ware multivariate Gaussian mixture model that models the datapoint representation, error-distance, and model prediction (\emph{e.g.} confidence scores in classification tasks). 
The observations of one datapoint from a model $\mathcal{M}$ include three components: $\{Z,E, \mathcal{Y}\}$, where $Z$ is an embedding representation, $\mathcal{Y}$ is predicted probabilities or confidence scores from the model, and error-distance $E$ is the distance between the one-hot tensor of the gold label $Y$ and $\mathcal{Y}$:
\begin{align}
E = Y - \mathcal{Y}
\end{align}

\begin{figure}[!ht]
    \centering
    \includegraphics[scale=0.4]{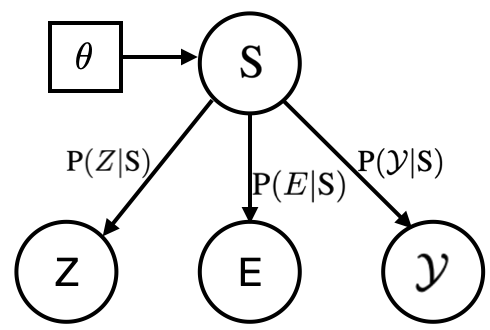}
    \caption{Model Structure}
    \label{fig:model}
    \vspace{-5pt}
\end{figure}

For each datapoint, $Z$ encodes the task-relevant semantic information;
$E$ encodes both label information and confidence information, which represents whether the prediction is wrong, to what extent it deviates from the gold label, and how much change is still required to make a correct prediction; $\mathcal{Y}$ encodes the confidence score, which is added to the model to control the weights of label information and of confidence information.\footnote{Future work can replace confidence with calibrated confidence to improve the model \cite[inter alia.]{calibrationyu, calineural, oodcali, kernelcali}, since calibrated confidence usually presents a better probability estimate of the likelihood for a datapoint to be categorized in some class.} We perform PCA on representations to filter out redundant information before applying the SDM.
Figure \ref{fig:model} illustrates the model structure. 

The generative story of the \SDM model is as follows: in order to generate all the observations of one datapoint, one slice $S_j$ is first drawn from a categorical distribution with parameters $\theta$. Then, the embedding $Z$, the error-distance $E$ and model prediction $\mathcal{Y}$ are drawn from slice-specific Gaussian distributions with parameters $\{\mu^{Z}_j, \Sigma^{Z}_j\}, \{\mu^E_j, \Sigma^E_j\}\ \text{and}\ \{\mu^{\mathcal{Y}}_j, \Sigma^\mathcal{Y}_j\}$, respectively.
\begin{align}
    S_j &\sim P(S;\theta)\nonumber \\
    Z|S_j &\sim \mathcal{N}(\mu^{Z}_j, \Sigma^{Z}_j)\nonumber \\
    E|S_j &\sim \mathcal{N}(\mu^E_j, \Sigma^E_j)\nonumber \\
    \mathcal{Y}|S_j &\sim \mathcal{N} (\mu^{\mathcal{Y}}_j, \Sigma^\mathcal{Y}_j)
\end{align}
For each datapoint $d$, the joint likelihood of slice $S_j$ and the observations of $d$ is a weighted product of the likelihoods from all distributions in the model with weights $\gamma,\lambda_E, \lambda_{\mathcal{Y}}$ on the Gaussians:
\begin{equation}
\small{
    L(d, S_j) = P(S_j)P(Z_d|S_j)^{\gamma}P(E_d|S_j)^{\lambda_E}P(\mathcal{Y}_d|S_j)^{\lambda_{\mathcal{Y}}}
}
\end{equation}
Given the joint likelihoods, the conditional probability of slice assignment $P(S_j|d)$ for the datapoints can be computed as:
\begin{equation}
P(S_j|d) = \frac{L(d, S_j)}{\Sigma_{j=1}^{j=k}L(d, S_j)}\propto L(d, S_j)
\end{equation}
Semantic information in the embedding, the error-distance, and the model predictions together determine the slice distribution. Thus datapoints that share some similar semantic features with the same gold label and similar model predictions are encouraged to be clustered into one slice. Given the joint likelihood, each slicing function $\psi_j$ is defined such that $\forall d\in \mathcal{D}, \psi_j(d) = 1$ if and only if:
\begin{equation}
\text{argmax}_iL(d, S_i) = j
\end{equation}

\subsection{Train}
The model parameters are estimated with Expectation-Maximization by maximizing the sum of log-likelihood of all datapoints $d\in\mathcal{D}$ in each slice $S_j$ for $j\in\{1, ..., k\}$:
\begin{equation}
    \mathcal{L}(\mathcal{D}) = \Sigma_{i=1}^{i=|\mathcal{D}|}\log\Sigma_{j=1}^{j=k} L(d_i, S_j)
\end{equation}
in which the assignment likelihood and the model parameters are estimated iteratively.
\SDM is tuned using the embeddings, error-distances, and confidence scores from the validation dataset of a task after $\mathcal{M}$ has been trained on the training dataset.

A slice $S_j$ is defined as an error slice, denoted as $S^e_j$ if the accuracy of $\{d\in\mathcal{D}|d\in S_j\}$ < $\delta$
for some threshold $\delta\in\mathbb{R}$. We call slicing functions corresponding to error slices as error slicing functions, denoted as $\psi_j^e$ corresponding to $S_j^e$.

\subsection{Inference}
For inference, we apply the tuned \SDM to test datasets $\mathcal{T}$ where gold labels are unknown to the model. Since gold label information is not available, the error-distance needs to be marginalized over potential label values. Thus the joint likelihood of a test datapoint $t\in \mathcal{T}$ and slice $S_j$ is computed as below, where $E'_t$ ranges over all possible $E$ values:
\vspace{-10pt}
\begin{align}
    L(t, S_j) =& P(S_j)P(Z_t|S_j)^{\gamma}(\Sigma_{E'_t}P(E_t'|S_j)^{\lambda_E})\nonumber \\
    &\cdot P(\mathcal{Y}_t|S_j)^{\lambda_{\mathcal{Y}}}
\end{align}
Then for each datapoint $t$, $\psi_j(t) = 1$ if and only if
\begin{equation}
\text{argmax}_iL(t, S_i) = j
\end{equation}

An unlabeled datapoint $t$ is determined to be error-prone if $\psi_j^e(t) = 1$ for some $j\in\{1, ..., k\}$.

\subsection{Comparison with DOMINO}

The difference between \SDM and DOMINO has notable empirical effects while theoretically nuanced. In \SDM, all distributions $Z|S_j, E|S_j, \mathcal{Y}|S_j$ are continuous and thus modeled by Gaussian distributions. This is enabled by converting the discrete $Y$ into a continuous $E$, which still preserves the label information.  While in DOMINO, only the distribution of $Z|S_j$ is modeled as Gaussian, while both $Y|S_j$ and $\mathcal{Y}|S_j$ are treated as categorical distributions because $Y$ is a discrete variable and $\mathcal{Y}$ is usually treated in the same manner as $Y$. Consequently, \SDM comprises an array of Gaussian distributions, whereas DOMINO combines Gaussian and categorical distributions. This subtlety results in different levels of empirical difficulty during hyperparameter searching: 
a model consisting only of Gaussian distributions allows a much larger range of effective hyperparameters that can achieve good performance in the evaluation benchmark across all three SDM facets, especially in the Discover and Improvement parts. Thus empirically \SDM is much easier to tune and can obtain better performance. More detailed experimental results and comparative analysis will be discussed in Section 5.

\section{\name Benchmark}

The \name Benchmark evaluates the performance of a tuned SDM. This section elaborates on the specifics of the three modules within the \name Benchmark: (1) the process of error-prone data point detection (Discover), (2) the manner in which explanations are delivered (Explain), (3) the approach to model improvement (Improvement), and the evaluation metrics for each module.

\subsection{Discover: Error-prone Datapoints Detection}
\label{discover}
In the Discover module, the objective is to ascertain if an SDM, after recognizing the error patterns present in the validation dataset, can accurately identify datapoints that are challenging for $\mathcal{M}$. As such, we deploy a tuned SDM on unlabeled datasets, anticipating it to correctly pinpoint error-prone datapoints. The details of this process are elaborated in the preceding Inference subsection. To evaluate its efficacy, we simply resort to determining whether the selected datapoints are indeed mispredicted by $\mathcal{M}$.

\subsection{Explain: Slice Feature Detection}
\label{explain}
In the Explain module, the objective is to make errors more interpretable as well as actionable. Towards this end, we find features that significantly correlate with an error slice as explanations. Such features can be surface string features such as specific tokens, linguistic features such as part-of-speech, and pragmatic indicators. Note that the Explain module seeks to interpret errors, which necessitates knowing which datapoints are indeed mispredicted. As such, this process is conducted on the validation dataset where \SDM is tuned on.

Table \ref{tab:feature example} displays some instances of systematic errors in the CoLA dataset\footnote{https://nyu-mll.github.io/CoLA/}, a dataset for the grammaticality judgment task, that are easily interpretable. Sentences 1 - 3 are incorrectly predicted due to inappropriate preposition usage. The grammatically correct version would be "It is Kim on whom Sandy relies." Similarly, sentences 4 - 6 are mispredicted due to incorrect usage of superlatives and the correct would be "That's the kindest answer that I ever heard."
\begin{table}[ht]
\resizebox{7.8cm}{!}{
    \centering
    \begin{tabular}{l|l|l|l}
    \toprule
    index & input sentence & pred & label \\
    \hline
       1 &  It is Kim on whom Sandy relies on. & T & F \\
        \hline
       2 &  It is Kim whom Sandy relies. & T & F \\
        \hline
       3 &  It is on Kim on whom Sandy relies. & T & F\\
       \hline 
       \hline 
       4 & That's the most kind answer that I ever heard. & T & F \\
        \hline
       5 &  That's a most kind answer that I ever heard. & T & F \\
        \hline
       6 & That's a kindest answer that I ever heard. & T & F \\
        \bottomrule
    \end{tabular}
}
    \caption{Examples on CoLA dataset}
    \label{tab:feature example}
\end{table}

In order to elucidate possible explanations for systematic errors, we have constructed a feature benchmark consisting of 38 unique features, denoted as $F$. Each feature is associated with a corresponding function, denoted as $f$. This benchmark facilitates the intrinsic evaluation of slices pinpointed by an SDM.

\begin{table}[ht]
\resizebox{7.8cm}{!}{
\begin{tabular}{ll}
      \toprule
      \textbf{Feature Type} & \textbf{Features}  \\
      \midrule
       \makecell[l]{surface string \\features}  & \makecell[l]{length, word frequency in corpus, foreign word} \\
      \hline
      \makecell[l]{syntactic \\features} & \makecell[l]{negation, reflexive, aspect, tense,\\ voice, comparison, echo question, multiple modal,\\ multiple prepositions, NP-subordinate clause,\\ quantifier, long-distance dependency,\\ tree depth, extra infinite with modal,\\ how-question, why-question,\\} \\
      \hline
      \makecell[l]{pragmatic\\ features} & \makecell[l]{age, gender, nationality, physical appearance, \\race/ethnicity, religion, social economic status,\\ sexual orientation, toxicity,\\ valency sentiment (positive/negative/neutral),\\ arousal (excited/calm/neural),\\ dominance (dominant/subordinate/neutral),\\ number of people, number of organization, \\number of location, number of money,\\ date, product,ordinal\_number} \\
       \bottomrule
    \end{tabular}
    }
    \caption{Linguistic feature benchmark}
    \label{tab:featurebenchark}
\end{table}

Table \ref{tab:featurebenchark} presents all features grouped into three types in the benchmark: surface string features, syntactic features, and pragmatic features. Surface string features include features that can be detected based on surface strings such as sentence length, word frequency in the corpus, and whether the sentence contains foreign words. Synthetic features require a dependency parser or a constituency parser to detect, such as negation, reflexive, aspect, and so on. Pragmatic features include age, gender, nationality of people mentioned in the sentence, etc. detected by models trained on the corresponding task. Table \ref{tab:feature_exp} uses examples in the CoLA dataset to illustrate some syntactic features. 
\begin{table}
\resizebox{8cm}{!}{
\begin{tabular}{l l}
    \toprule
    \textbf{Feature} & \textbf{Explanation}\\
    \hline
     Reflexive & \makecell[l]{sentences containing reflexives (myself/themselves/\\each other and \emph{etc}) such as ``*John believes that \\Mary saw himself.''} \\
     \hline
     Aspect & \makecell[l]{sentences using perfect aspect such as\\ ``The men would have been all working.''}\\
     \hline
     Tense & sentence using past tense \\
     \hline
     Voice & \makecell[l]{sentences using passive voice such as \\``A mile to work was run by him.''}\\
     \hline
     Comparison & \makecell[l]{sentences containing comparative or superlative\\ such as ``He's more reliable a man.''} \\
     \hline
     Multiple Modal & \makecell[l]{sentences that wrongly contain multiple modals \\ such as ``*Kim must will bake a cake.''} \\
     \hline
     \makecell[l]{Multiple \\preposition} & \makecell[l]{sentences with many prepositions such as\\ ``This girl \textbf{in} the red coat will put a picture of Bill\\ \textbf{in} the mailbox and \textbf{on} your desk before tomorrow.''}\\
     \hline
     \makecell[l]{NP-subordinate\\clause} & \makecell[l]{sentences containing NP-subordinate clauses \\such as ``This is the book that we had read.''}\\
     \hline
     Quantifier & \makecell[l]{sentences containing numerals, any, all, most \\and \emph{etc} such as ``*Almost any owl hunts mice.''}\\
     \hline
     \makecell[l]{Long distance \\dependency} & \makecell[l]{sentences containing long syntactic dependencies \\such as ``The video which I thought John told us \\you recommended was really terrific.''}\\
     \hline
     Tree depth & \makecell[l]{sentences whose largest tree depths in their \\ constituency parse trees are exceptionally large such as \\``The pen of the girl's father's uncle's wife is beautiful''.}\\
     \hline
     \makecell[l]{Extra infinite \\after modal} & \makecell[l]{sentences that wrongly contain ``to'' after modals\\ such as ``*John can to kick the ball.''} \\
     \bottomrule
\end{tabular}
}
\caption{Syntactic feature examples}
\label{tab:feature_exp}
\end{table}

Each feature $F$ corresponds to a feature function $f$: if $F$ is binary such as negation and echo question, then $f$ is a characteristic function such that $f(\text{sentence}) = 1$ indicates that the sentence contains the feature; if $F$ is non-binary such as multiple-preposition and long-distance dependency, then $f(\text{sentence}) = d\in\mathbb{R}$ indicating that the sentence has $d$-degree of the feature.

To evaluate whether an SDM is able to group datapoints sharing the same error-correlating features, we design two feature discovery tasks: Synthetic Feature Detection and Real Dataset Feature Detection, evaluating whether an SDM is able to group datapoints sharing the same error-correlating features. 

The first task evaluates the feature discovery capability by using synthetic datasets where each dataset contains one gold error-correlated feature. A synthetic dataset with a feature $F$ is generated by mixing a set of wrongly predicted datapoints featuring $F$:
$D_{target} = \{d\in \mathcal{D}|\mathcal{M}(d) \neq \text{label}(d) \text{ and } f(d) = 1\}$ (assuming $f$ is a characteristic function here) and a set of randomly selected datapoints from the original dataset with the same number. Then we fit an SDM on the synthetic dataset to see how many target datapoints in $D_{target}$ can be grouped into error slices and then we can compute recall, precision, and F1.

The second task is to detect features in real datasets, which also characterizes how SDM can be utilized for general model analysis.
For each datapoint, we apply all feature functions to find out the set of features that it exhibits. 
Then for each error slice, 
we leverage significance testing to analyze which features are distributed significantly differently between in-slice and out-of-slice data. For each feature's in-slice and out-of-slice distributions, if the $p$-value from a permutation test is smaller than 0.05 and the mean of the in-slice distribution is larger than that of the out-of-slice distribution (as their occurrences usually complicate sentence structures), this feature is strongly correlated with erroneous predictions. 

Both tasks aim at finding the error-correlated features. These interpretable features describe these datapoints and provide insights into the behaviors of current models.

\subsection{Improve: Downstream Tasks}
\label{improvement}

The final module of the benchmark assesses the SDM's capacity to enhance model performance. Three automated improvement methods are utilized in this module: selective prediction, flipping, and active learning. When these techniques are deployed on a tuned SDM, they can boost model performance if the SDM can identify an ample amount of informative error patterns and error-prone datapoints. Consequently, these methods serve a dual function - demonstrating the feasibility of automated improvements using SDM and evaluating the SDM.

\subsubsection{Selective Prediction}
The selective prediction task aims at pointing out which datapoints are error-prone in a given unlabeled dataset $\mathcal{T}$ and rejecting them from being evaluated. 
An SDM predicts a datapoint $t$ to be error-prone if $t\in \mathcal{E}$ where $\mathcal{E} = \{t\in \mathcal{T}\mid \psi_{j}^e(t) = 1\text{ where }j\leq k\}$ with $\psi_{j}^e$ being some error slicing function. It reorders each $t$ based on error probability $P(e=1|t)$ of $t$ defined as below:
\vspace{-10pt}
\begin{multline}
    P(e=1|t) = \Sigma_{S^e \in S_*^e} \frac{L(t, S^e)}{\Sigma_{j=1}^{j=k} L(t, S_j)}. \\ 
    \text{where } S_*^e \text{ is the set of all error slices }
\end{multline}
It refrains from evaluating these datapoints one by one to demonstrate the change of efficacy of the remaining datapoints. The more efficacy increases, the better this task is fulfilled. 

\subsubsection{Flipping}

Flipping is a task to directly improve model performance by flipping the prediction of error-prone datapoints in an unlabeled dataset. If the dataset is binary, flipping changes its prediction from 1 to 0 or 0 to 1; if the dataset is multi-labeled, we select a label to flip the predicted label to. 

For multi-labeled datasets, for each error-prone datapoint $t$, we select the new label as follows: if the confidence score of $t$ is below some threshold and $\psi_j^e(t) = 1$ for some $j$, we find the majority of gold label $l$ in $S^e$ in validation dataset and flip $t$'s prediction to $l$; if the confidence of $t$ is above the threshold, the predicted label remains the same. The confidence threshold is found with 10$\%$ of the validation dataset used to train the SDM.

For the confidence baseline, the label is flipped to the next confident label in the corresponding error slice.

In flipping, the predicted error-prone datapoints $t$ are also flipped one by one ordered by $P(e=1|t)$ as in the selective prediction task.

\subsubsection{Active Learning}
Active learning is an interactive learning algorithm that proactively selects examples to be labeled next from the pool of unlabeled data. 
Error-prone datapoints are also points with potential bias and training with them should promote time and data efficiency. Thus if an SDM can accurately select enough error-prone datapoints, simulating active learning in training time will help the model learn faster in training.
The active learning simulation in \name is implemented as follows:
\textbf{Step 1}: divide the whole training dataset into a small training seed set and an extra training data pool from which more training datapoints can be selected to train the model on.
\textbf{Step 2}: fit an SDM on the validation dataset and select error-prone datapoints from the extra training data pool without using label information to replicate real-time scenario.
\textbf{Step 3}: create a new training dataset combining original training data + selected training data and remove the selected datapoints from the extra training data pool.
\textbf{Step 4}: retrain the model on the new training dataset.
Repeat steps 2-4 until the model converges on the validation dataset.

\section{Experiment Result}
This section presents experiment results for all three modules on \SDM. It illustrates how this benchmark should be used and also demonstrate that \SDM is able to cluster error datapoints with similar features and detect error-prone datapoints accurately. 

We apply \name on a variety of datasets in GLUE benchmark \cite{glue} and Kaggle dataset Jigsaw\footnote{https://www.kaggle.com/competitions/jigsaw-unintended-bias-in-toxicity-classification/overview/evaluation}: CoLA, QNLI, QQP, SST-2, MNLI, SST-5, Jigsaw-gender, Jigsaw-racial, Jigsaw-religion.  Since GLUE test dataset labels are not publicly available, we split the original training dataset into training and validation, and treat the original validation dataset as a test dataset.
For each dataset, we train three models based on three widely used models: BERT-large, RoBERTa-large, and ELECTRA-large-discriminator with the following hyperparameters: $\{$batch size = 32, learning rate = 1e-4, warm up proportion = 0.1, epochs = 10, gradient clip = 1.0, dropout rate = 0.1$\}$. All models are trained on one A5000 GPU. To evaluate the performance of \name, we apply \SDM on each of the trained models and evaluate on results from these models. 

\subsection{Discover: Experiment Result}
In the Discover module, the evaluation of an SDM's performance with respect to any model $\mathcal{M}$ is achieved by assessing its efficacy in identifying error-prone datapoints, that is, determining whether these points are indeed mispredicted by $\mathcal{M}$. We compare the performance of \SDM with DOMINO which is the current SOTA slice detection model, confidence thresholding, and random sampling. 

\begin{table*}[t!]
    \centering
    \resizebox{17cm}{!}{
    \begin{tabular}{c|cccccc|cccccc|cccccc}
    \toprule
        \textbf{NLP model} & \multicolumn{6}{c|}{\textbf{BERT}} & \multicolumn{6}{c|}{\textbf{RoBERTa}} & \multicolumn{6}{c}{\textbf{ELECTRA}}\\
        \hline
        \textbf{Method} & \multicolumn{2}{c}{\textbf{\SDM}} & \multicolumn{2}{c}{\textbf{DOMINO}} & \multicolumn{1}{c}{\textbf{conf}} &\multicolumn{1}{c|}{\textbf{rand}} & \multicolumn{2}{c}{\textbf{\SDM}} &
        \multicolumn{2}{c}{\textbf{DOMINO}} &
        \multicolumn{1}{c}{\textbf{conf}}  &\multicolumn{1}{c|}{\textbf{rand}} & \multicolumn{2}{c}{\textbf{\SDM}} &
        \multicolumn{2}{c}{\textbf{DOMINO}} &
        \multicolumn{1}{c}{\textbf{conf}}  &\multicolumn{1}{c}{\textbf{rand}} \\
        \hline
        \textbf{Metric} & \textbf{num} & \textbf{eff} & \textbf{num} & \textbf{eff} & \textbf{eff} & \textbf{eff} & \textbf{num} & \textbf{eff} & \textbf{num} & \textbf{eff} &  \textbf{eff} & \textbf{eff} & \textbf{num} & \textbf{eff} & \textbf{num} & \textbf{eff} &  \textbf{eff} & \textbf{eff} \\
        \hline
        CoLA & 38 & \textbf{63.16} & 74 & 33.78 & 55.26 & 21.86 & 96 & \textbf{60.42} & 52 & 32.69 & 45.83 & 28.67 & 6 & \textbf{50.00} & 61 & 29.51 & \textbf{50.00} & 13.82\\
        QNLI & 313 & \textbf{60.06} & 204 & 41.48 & 53.04 & 26.10 & 348 & \textbf{54.70} & 464 & 39.01 & 45.12 & 26.67 & 85 & \textbf{57.65} & 103 & 35.92 & 48.24 & 8.82\\
        QQP & 1728 & \textbf{51.45} & 2349 & 42.32 & 47.23 & 23.56 & 652 & 48.00 & 1802 & 41.18 & \textbf{50.92} & 23.57 & 152 & \textbf{59.21} & 3159 & 35.84 & 43.42 & 19.38\\
        SST-2 & 27 & \textbf{48.15} & 35 & 37.14 & 40.74 & 21.90 & 24 & \textbf{54.37} & 33 & 42.42 & 50.00 & 20.87 & 36 & \textbf{47.22} & 29 & 44.83 & 41.67 & 12.01\\
        MNLI & 530 & \textbf{61.13} & 465 & 55.48 & 60.94 & 35.27 & 441 & 61.68 & 975 & 47.08 & \textbf{62.81} & 37.23 & 110 & \textbf{57.27} & 125 & 31.20 & \textbf{57.27} & 13.21\\
        SST-5 & 667 & \textbf{59.52} & 783 & 55.56 & 56.97 & 51.45 & 705 & 51.91 & 789 & 48.92 & \textbf{52.77} & 46.45 & 119 & \textbf{60.50} & 813 & 49.08 & 53.78 & 43.17\\
        J-gender & 1160 & 65.43 & 1501  & \textbf{70.09} & 50.17 & 43.47 & 1334 & \textbf{65.42} & 1592 & 57.67 & 52.61 & 44.98 & 304 & 32.57 & 132 & 31.06 & \textbf{43.75} & 12.39\\
        J-racial & 1049 & \textbf{73.59} & 1247 & 66.73 & 48.62 & 39.59 & 1254 & \textbf{72.81} & 1554 & 62.36 & 47.60 & 41.55 & 356 & 43.82 & 166 & 35.54 & \textbf{44.66} & 16.01\\
        J-religion & 135 & \textbf{87.41} & 77 & 53.25 & 48.62 & 17.43 & 142 & 43.66 & 82 & \textbf{48.78} & 43.66 & 14.39 & 166 & \textbf{78.92} & 121 & 46.28 & 54.22 & 24.39\\
        \hline 
        \hline
        average & N/A & \textbf{63.32} & N/A & 50.65 & 51.28 & 31.18 & N/A & \textbf{57.00} & N/A & 46.68 & 50.15 & 31.60 & N/A & \textbf{51.16} & N/A & 37.70 & 47.28 & 18.13\\
        \bottomrule
    \end{tabular}
    }
    \vspace{-5pt}
    \caption{Efficacy of predicted error-prone datapoints}
    \label{tab:selected_wrong_performance}
    \vspace{-15pt}
\end{table*}

The hyperparameters for \SDM in the Discover module are $\{\gamma = 0.15, \lambda_{E} = 0.1, \lambda_{\mathcal{Y}} = 1$, PCA dimension = 128, number of slices = 128$\}$ for all datasets for all models BERT, RoBERTa, and ELECTRA. For DOMINO, we manually tune their hyperparameters for the best performance.\footnote{The set of hyperparameters is $\gamma = 1, \lambda_{Y} = 10, \lambda_{\mathcal{Y}} = 40$ after manual tuning. This is different from the default hyperparameters provided in \citet{domino}: $\gamma = 1, \lambda_{Y} = 10, \lambda_{\mathcal{Y}} = 10$, tuned for feature detection on CV.} Both sets of hyperparameters are tuned only on held-out sets in CoLA using BERT models.


Table \ref{tab:selected_wrong_performance} reports the test dataset results: (1) the number of error-prone datapoints found and (2) efficacy, which is defined by
\begin{equation}
    \frac{\lvert \{t\in \mathcal{E}_{SDM}| \mathcal{M}(t) \neq \text{label}(t)\}\rvert}{\lvert\mathcal{E}_{SDM}\rvert}
\end{equation}
where $\mathcal{E}_{SDM}$ is the set of predicted error-prone datapoints predicted by the SDM.

The efficacy for confidence baseline is computed on $\lvert\mathcal{E}_\SDM\rvert$ lowest confident datapoints; the efficacy for random baseline is computed based on $\lvert\mathcal{E}_\SDM\rvert$ randomly sampled datapoints.\footnote{We do not compute confidence baseline and random sampling baseline based on the number of error-prone datapoints discovered by DOMINO because, in all GLUE datasets, DOMINO's efficacy is lower than confidence baseline efficacy.} 
Seen from Table. \ref{tab:selected_wrong_performance}, the efficacies of \SDM are almost always much higher than other baselines and higher than 50.00$\%$, indicating that it is effective in discovering datapoints that will be mispredicted by $\mathcal{M}$. 

Among the three types of models BERT-large, RoBERTa-large and ELECTRA-large-discriminator, \SDM performs much better in the former two. It could be because ELECTRA-large-discriminator already performs very well in all the nine datasets and \SDM is not able to witness enough mispredicted datapoints in validation datasets in tuning time to generalize to test datapoints during inference.

\subsubsection{Model Structure Ablation}
We study the model structure based on the efficacy performance on CoLA and QNLI in Table \ref{tab:structure_abla}. We compare models with (1) only $\mathcal{Y}$ edge (\SDM-$\mathcal{Y}$) (2) both $E$ and $\mathcal{Y}$ edge (\SDM-$E,\mathcal{Y}$) and (3) all three edges (\SDM). 

First, we notice that \SDM-$\mathcal{\mathcal{Y}}$ detects error-prone datapoints more accurately than confidence baseline, 
indicating that selecting error slices with a certain range of confidence scores validated by validation datasets is better than directly choosing datapoints with the lowest confidence scores throughout the whole dataset, as efficacy is not always directly related to confidence score. \SDM-$E,\mathcal{Y}$ calibrates confidence scores, which performs more accurately. \SDM leveraging representation information selects error-prone datapoints most accurately, indicating that semantic information provides additional clues on difficulty levels of datapoints for a given model, which contributes to error detection.

\begin{table}[ht]
\centering
\resizebox{7.8cm}{!}{
\begin{tabular}{l cccc}
      \toprule
       & \textbf{dataset} & \textbf{number} & \textbf{efficacy} & \textbf{confidence} \\
       \midrule
      \multirow{2}{*}{\SDM-$\mathcal{Y}$} & CoLA & 78 & 44.87 & 41.03\\
       & QNLI & 306 & 57.84 & 53.99\\
      \hline
      \multirow{2}{*}{\SDM-$E,\mathcal{Y}$} & CoLA & 73 & 46.58 & 43.84\\
       & QNLI & 285 & 57.95 & 54.04\\
      \hline
      \multirow{2}{*}{\SDM} & CoLA & 38 & 63.16 & 55.26 \\
       & QNLI & 313 & 60.06 & 53.04 \\
       \bottomrule
    \end{tabular}
    }
    \caption{Ablation study on model structure}
    \label{tab:structure_abla}
\end{table}

\subsubsection{Validation Size Ablation}
\begin{figure}[!h]
    \centering
    \includegraphics[scale=0.36]{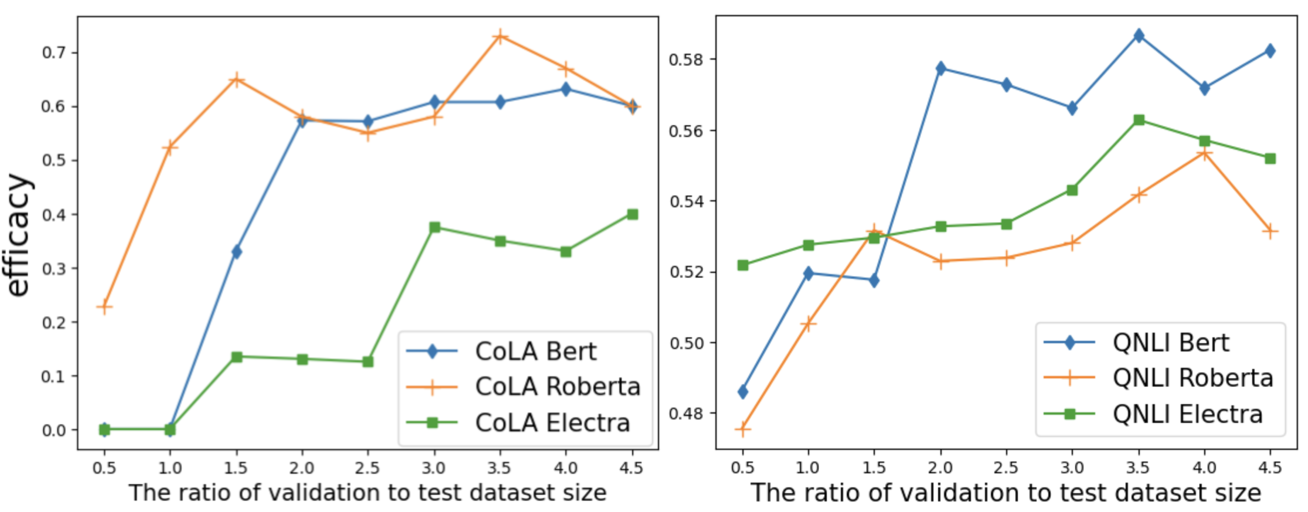}
    \caption{Efficacy with different sizes of validation}
    \label{fig:validation_size}
\end{figure}

We investigate how the size of the validation dataset, on which \SDM is tuned, impacts the efficacy of the model. Ideally, the larger the validation size, the more error patterns it potentially covers, and the better the result.
Figure. \ref{fig:validation_size} uses the CoLA and QNLI datasets as examples to present the correlation between validation dataset size and the model's efficacy. The x-axis is the ratio of the validation dataset size to the test dataset size, and the y-axis is the efficacy. As a reference, the test dataset for CoLA contains 1043 datapoints, while that of QNLI contains 5463 datapoints. From the figures, we can draw the following conclusions: (1) In general, the larger the validation dataset used to train the \SDM, the higher the model's efficacy. (2) If the validation dataset is smaller than the test dataset, \SDM's performance decreases a lot, especially for CoLA, which has a small test dataset. Based on this validation result, to ensure adequate coverage of error patterns, we recommend that the validation dataset be at least twice the size of the test dataset.

\subsubsection{Hyperparameter Sensitivity}
We explore different settings of \SDM and test functions of the following hyper-parameters: weights ($\lambda_E$ and $\gamma$ with $\lambda_\mathcal{Y}$ fixed) and PCA dimension. We conduct experiments on the BERT-based model on CoLA and QNLI datasets. 

\paragraph{Weights}
We test different weights with $\lambda_E\in\{0.01, 0.05, 0.2, 0.5, 1\}$ and $\gamma\in\{0.01, 0.05, 0.2, 0.5, 1\}$ separately.
\begin{table}[!ht]
    \centering
    \resizebox{8cm}{!}{
    \begin{tabular}{c|ccc|ccc}
        \toprule
         & \multicolumn{3}{c|}{\textbf{CoLA}} & \multicolumn{3}{c}{\textbf{QNLI}} \\
         \hline
        \textbf{$\lambda_E$} & \textbf{number} & \textbf{efficacy} & \textbf{conf} & \textbf{number} & \textbf{efficacy} & \textbf{conf} \\
         \hline
        0.01 &  38 & \textbf{63.16} & 55.26 & 285 & \textbf{58.60} & 54.04\\
        \hline
        0.05 &  40 & \textbf{55.00} & 55.00 & 266 & \textbf{59.02} & 53.38\\
        \hline
        0.2 &  33 & \textbf{60.61} & 54.54 & 346 & \textbf{58.95} & 51.45 \\
        \hline
        0.5 & 117 & 40.17 & \textbf{43.59} & 1732 & 36.54 & \textbf{42.67}\\
        \hline
        1 & 120 & 40.00 & \textbf{50.00} & 1705 & 37.83 & \textbf{43.17}\\
        \hline
        \hline
        \textbf{$\gamma$} & \textbf{number} & \textbf{efficacy} & \textbf{conf} & \textbf{number} & \textbf{efficacy} & \textbf{conf} \\
        \hline
        0.01 & 65 & 37.91 & \textbf{44.62} & 233 & \textbf{54.09} & 53.22\\
        \hline
        0.05 & 39 & 46.25 & \textbf{53.85} & 232 & \textbf{62.93} & 53.02\\
        \hline
        0.2 & 58 & \textbf{50.00} & 48.28 & 312 & \textbf{58.65} & 52.88\\
        \hline
        0.5 & 28 & \textbf{42.87} & 42.87 & 315 & \textbf{59.37} & 53.33\\
        \hline
        1 & 1 & 0.00 & \textbf{100.00} & 111 & \textbf{58.56} & 51.35\\
        \bottomrule
    \end{tabular}
    }
    \caption{Ablation study on the value of $\lambda_{E}$ and $\gamma$}
    \label{tab:gamma_D}
\end{table}

In Table \ref{tab:gamma_D}: 
(1) For $\lambda_{E}$, efficacy is high with values smaller than 0.5 but decreases with large values. With large $\lambda_{E}$, \SDM overfits on the validation dataset because there is a discrepancy between the training and the testing modeling scheme: when fitting an SDM on the validation dataset, the model leverages all information of input representations, error-distance, and model predictions; while in the test stage, it does not have access to the ground truth information. Thus focusing on error-distance information when tuning on validation misleads the model and negatively impacts the performance of the test dataset. 
(2) For $\gamma$, it impacts negatively on efficacy with both small and large values. Large values are harmful because semantic representation does not have a straightforward relationship with prediction results for any given model $\mathcal{M}$. Thus focusing mainly on semantic feature information while neglecting label and prediction information encourages a flatter accuracy distribution over slices, and thus it is more difficult to find high-quality error slices in validation to fit on the test dataset. 
Small values hurt performance may be because they render input representation information to be noise to \SDM and thus impact the performance negatively.

\paragraph{PCA dimensions}
We test PCA dimension = 32, 64, 128, 256, and 1024 (without PCA dimension reduction) under different weights of the embedding. The results are presented in Table \ref{tab:pca_ablation}.
\begin{table}[!ht]
    \centering
     \resizebox{8cm}{!}{
    \begin{tabular}{cc|ccc|ccc}
    \toprule
        & & \multicolumn{3}{c|}{\textbf{CoLA}} & \multicolumn{3}{c}{\textbf{QNLI}} \\
        \hline
        \textbf{PCA} & \textbf{$\gamma$} & \textbf{number} & \textbf{efficacy} & \textbf{conf} & \textbf{number} & \textbf{efficacy} & \textbf{conf} \\
        \hline
        32 & 0.15 & 55 & \textbf{52.72} & 50.91 & 289 & \textbf{56.04} & 54.03\\
        \hline
        64 & 0.15 & 78 & \textbf{43.59} & 41.03 & 290 & \textbf{60.00} & 53.79 \\
        \hline
        256 & 0.15 & 4 & 25.00 & \textbf{50.00} & 297 & \textbf{55.89} & 54.21\\
        \hline
        1024 & 0.15 & 1 & 0.00 & \textbf{100.00} & 229 & \textbf{53.95} & 53.20\\
        \hline
        \hline
        32 & 0.1 & 53 & 37.74 & \textbf{50.94} & 232 & \textbf{58.18} & 53.00 \\
        \hline 
        64 & 0.1 & 23 & \textbf{65.32} & 60.87 & 295 & \textbf{60.00} & 53.90\\
        \hline
        256 & 0.1 & 49 & \textbf{55.11} & 48.98 & 323 & \textbf{57.59} & 52.60 \\
        \hline
        1024 & 0.1 & 4 & 25.00 & \textbf{50.00} & 229 & 52.39 & \textbf{53.20}\\
        \hline
        \hline
        32 & 0.05 & 85 & \textbf{45.88} & 41.18 & 301 & \textbf{54.17} & 53.82 \\
        \hline
        64 & 0.05 & 66 & \textbf{50.00} & 45.45 & 301 & \textbf{60.13} & 53.82\\
        \hline
        256 & 0.05 & 39 & \textbf{56.41} & 53.85 & 298 & \textbf{58.05} & 54.03 \\
        \hline
        1024 & 0.05 & 1 & 0.00 & \textbf{100.00} & 235 & 52.15 & \textbf{53.61}\\
        \bottomrule
    \end{tabular}
    }
    \caption{Ablation study on PCA dimensions }
    \label{tab:pca_ablation}
\end{table}

For all three $\gamma$ values, PCA dimensions 64 and 256 work well. Embeddings without PCA dimension reduction perform much worse: In the CoLA dataset, it almost completely fails \SDM and \SDM can discover almost no error-prone datapoints; In QNLI dataset, the model performs non-trivial efficacy results but is still worse than when using other PCA dimensions. Thus in general, we recommend removing redundant information and noise by PCA dimension reduction.

Furthermore, we notice that the models using small dimensions (32) tend to work better under relatively large $\gamma$ values than small $\gamma$ values; the model using large dimensions (256) performs better with small $\gamma$ values than with large $\gamma$ values. Thus the PCA dimension should be chosen inversely to $\gamma$.


\subsection{Explain: Experiment Result}
Synthetic Dataset Feature Discovery and Real Dataset Feature Discovery evaluate how reliably an SDM can find feature explanations for errors. In these tasks, \name explains slices discovered in the validation dataset instead of the test dataset because it is expected to explain why the models fail on some data points which require access to gold labels. Therefore a different set of hyperparameters is required: $\{\gamma = 0.15, \lambda_E = 1, \lambda_\mathcal{Y} = 0.1\}$. Both experiment results demonstrate that \SDM performs better than DOMINO.

\paragraph{Synthetic Dataset Feature Discovery}
\begin{table}[t!]
    \centering
    \resizebox{8cm}{!}{
    \begin{tabular}{c c c c c}
    \toprule
        \textbf{dataset} & \textbf{model} & \textbf{avg. precision} & \textbf{avg. recall} & \textbf{avg. F1}\\
        \hline
        \multirow{2}{*}{CoLA} & \SDM & 26.04 & 95.17 &  \textbf{40.89} \\
         & DOMINO & 25.98 & 83.42 & 39.62\\
        \hline
        \multirow{2}{*}{QNLI} & \SDM & 7.83 & 25.12 & \textbf{11.94} \\
         & DOMINO & 7.32 & 17.77 & 10.37 \\
        \hline
        \multirow{2}{*}{QQP} & \SDM & 7.44 & 29.33 & \textbf{11.87} \\
         & DOMINO & 7.94 & 20.73 & 11.48 \\
        \hline
        \multirow{2}{*}{SST-2} & \SDM & 8.68 & 12.31 & \textbf{10.18} \\
         & DOMINO & 8.06 & 10.5 & 9.12\\
        \hline
        \multirow{2}{*}{MNLI} & \SDM & 7.59 & 34.58 & \textbf{12.45}\\
         & DOMINO & 7.88 & 14.17 & 10.12\\
        \hline
        \multirow{2}{*}{SST-5} & \SDM & 7.43 & 56.71 & \textbf{13.14} \\
         & DOMINO & 7.15 & 48.22 & 12.45 \\
        \hline  
        \multirow{2}{*}{Jigsaw-gender} & \SDM & 52.38 & 95.12 & 67.56 \\
         & DOMINO & 51.96 & 99.45 & \textbf{68.26} \\
        \hline
        \multirow{2}{*}{Jigsaw-racial} & \SDM & 29.49 & 88.36 & 44.22 \\
         & DOMINO & 29.30 & 99.30 & \textbf{45.24}\\
        \hline
        \multirow{2}{*}{Jigsaw-religion} & \SDM & 27.67 & 95.87 & \textbf{42.96} \\
         & DOMINO & 26.18 & 88.12  & 40.37\\
        \hline 
        \hline 
        \multirow{2}{*}{cross-dataset} & \SDM & \textbf{19.40} & \textbf{59.13} & \textbf{29.22}\\
         & DOMINO & 19.08 & 53.52 & 28.13 \\
        \bottomrule
    \end{tabular}
    }
    \caption{Synthetic feature detection result}
    \label{tab:inpaper_synthetic_exp}
    \vspace{-10pt}
\end{table}
We perform this experiment on features with a relatively large number of wrongly predicted datapoints: $\{$length, negation, reflexive, comparison, NP\_subordinate, multiple\_preposition, quantifier, tree\_depth, long-distance$\}$ for GLUE datasets and $\{$female, male, Asian, Black, White, Latino, Atheist, Buddhist, Christian, Hindu, Jewish, Muslim$\}$ for Jigsaw datasets. We compare \SDM results with DOMINO results.

Table \ref{tab:inpaper_synthetic_exp} presents cross-feature average precision, recall, and F1 for each dataset. 
In general \SDM performs better than DOMINO except in SST-2 where the average F1 of DOMINO result is +0.02 higher than that of \SDM. \SDM performs better in recall in all cases and better in precision in some cases. The last two rows of the table show the cross-dataset average precision, recall, and F1. We notice that \SDM performs better than DOMINO on all metrics, especially recall.

\paragraph{Hyperparameter Ablation}
We study the effect of hyperparameters in feature detection-related tasks on the CoLA dataset, which is a dataset focusing on grammaticality. Results on the effects of $\gamma$ and $\lambda_\mathcal{Y}$ with fixed $\lambda_E$ are presented  in Table.\ref{tab:inpaper_synthetic_ablation}. We noticed that large $\gamma$ improves precision but decreases the recall while large $\lambda_\mathcal{Y}$ brings the reverse effect. 
$\gamma = 1$ and $\lambda_\mathcal{Y} = 1$ have low recall because the former fails to detect feature Comparison and the latter fails to detect feature NP$\_$sub.
\begin{table}[!ht]
    \centering
    \resizebox{6cm}{!}{
    \begin{tabular}{cccc}
    \toprule
        \textbf{hyper-parameter} & \textbf{avg. precision} & \textbf{avg. recall} & \textbf{avg. F1}\\
        \hline
         & 26.04 & 95.17 & \textbf{40.89}\\
        \hline
        \hline
        $\gamma = 0.5$ & 21.78 & 74.88 & 33.72 \\
        \hline
        $\gamma = 1$ & \textbf{29.91} & 56.11 & 35.94\\
        \hline
        \hline
        $\lambda_\mathcal{Y} = 0$ & 25.19 & 94.89 & 35.00\\
        \hline
        $\lambda_\mathcal{Y} = 0.5$ & 24.90 & \textbf{97.13} & 38.28 \\
        \hline
        $\lambda_\mathcal{Y} = 1$ & 23.16 & 84.34 & 34.02\\
        \bottomrule
    \end{tabular}
    }
    \caption{Ablation study on synthetic feature detection}
    \label{tab:inpaper_synthetic_ablation}
\end{table}

\paragraph{Real Dataset Feature Detection}

\begin{table}[!ht]
    \centering
    \resizebox{8cm}{!}{
    \begin{tabular}{c  c  c  c  c  c}
    \toprule
        \textbf{dataset} & \textbf{model} & \textbf{feature prop} & \textbf{V-measure} & \textbf{Homo} & \textbf{Comp} \\
        \midrule
        \multirow{4}{*}{CoLA} & \SDM & 81.25 & 22.58 & 13.18 & 78.80 \\
         & DOMINO & 68.75 & 15.98 & 7.77 & 74.18 \\
         & \SDM-$Z$ &56.25 & 24.57 & 18.55 & 84.41 \\ 
         & \SDM-$E$ & 56.25 & 9.48 & 5.37 & 68.47 \\
        \midrule
        \multirow{4}{*}{QNLI} & \SDM & 50.00 & 7.11 & 3.78 & 60.12\\
         & DOMINO & 75.00 & 5.77 & 2.88 & 67.07 \\
         & \SDM-$Z$ &12.50 & 5.35 & 2.96 & 61.00\\
         & \SDM-$E$ &50.00 & 6.99 & 3.81 & 57.48\\
        \midrule
        \multirow{4}{*}{QQP} & \SDM & 50.00 & 7.23 & 3.87 & 68.31\\
         & DOMINO & 56.25 & 16.13 & 9.94 & 67.45\\
         & \SDM-$Z$ &12.50 & 15.4 & 9.30 & 48.41\\
         & \SDM-$E$ &50.00 & 10.68 & 3.19 & 76.15\\
        \hline
        \multirow{4}{*}{SST-2} & \SDM & 50.00 & 4.68 & 2.44 & 57.34\\
         & DOMINO & 62.50 & 4.01 & 1.97 & 69.46\\
         & \SDM-$Z$ &31.25 & 9.61 & 4.08 & 78.61\\
         & \SDM-$E$ &50.00 & 4.08 & 2.29 & 47.16\\
        \hline
        \multirow{4}{*}{MNLI} & \SDM & 68.75 & 17.00 & 10.12 & 53.22\\
         & DOMINO & 62.50 & 16.09 & 11.03 & 54.56\\
         & \SDM-$Z$ &50.00 & 18.51 & 13.89 & 46.18\\
         & \SDM-$E$ &50.00 & 15.64 & 9.42 & 65.59 \\
        \hline
        \multirow{4}{*}{SST-5} & \SDM & 81.25 & 23.52 & 14.23 & 67.78\\
         & DOMINO & 75.00 & 19.50 & 13.64 & 68.80\\
         & \SDM-$Z$ &31.25 & 23.30 & 43.79 & 31.95\\
         & \SDM-$E$ &62.50 & 21.30 & 16.62 & 65.33\\
        \hline
        \multirow{4}{*}{J-gender} & \SDM & 100.00 & 35.94 & 26.88 & 54.23\\
         & DOMINO & 100.00 & 36.85 & 26.73 & 54.42\\
         & \SDM-$Z$ &100.00 & 34.99 & 27.57 & 47.87\\
         & \SDM-$E$ &100.00 & 42.3 & 31.32 & 52.39 \\
        \hline
        \multirow{4}{*}{J-racial} & \SDM & 75.00 & 22.71 & 16.54 & 36.20\\
         & DOMINO & 100.00 & 33.44 & 22.03 & 63.25 \\
         & \SDM-$Z$ &100.00 & 27.09 & 19.67 & 53.88 \\
         & \SDM-$E$ &50.00 & 37.49 & 31.26 & 47.02\\
        \hline
        \multirow{4}{*}{J-religion} & \SDM & 100.00 & 46.13 & 41.23 & 52.35\\
         & DOMINO & 50.00 & 33.98 & 26.72 & 49.88\\
         & \SDM-$Z$ &83.33 & 38.53 & 28.30 & 49.39\\
         & \SDM-$E$ &83.33 & 23.62 & 23.49 & 28.79\\
        \hline
        \hline 
        \multirow{4}{*}{ave. weighted} & \SDM & --- & \textbf{21.74} & \textbf{14.69} & \textbf{45.47}\\
         & DOMINO & --- & 20.30 & 10.55 & 43.25 \\
         & \SDM-$Z$ &--- & 14.12 & 12.03 & 30.11 \\
         & \SDM-$E$ &--- & 13.69 & 10.45 & 32.98 \\
        \bottomrule
    \end{tabular}
    }
    \caption{Real datasets feature detection results}
    \label{tab:feature_detection}
    \vspace{-5pt}
\end{table}

In Table \ref{tab:feature_detection}, we report the error slice feature detection results with surface and syntactic features on GLUE datasets and pragmatic features on Jigsaw datasets. We compare with DOMINO model results, \SDM using only semantic embedding information (\SDM-$Z$) and that using only error-distance information (\SDM-$E$). 

\begin{table*}[t!]
    \centering
    \resizebox{17cm}{!}{
    \begin{tabular}{c|cccccc|cccccc|cccccc}
    \toprule
        \textbf{NLP model} & \multicolumn{6}{c|}{\textbf{BERT}} & \multicolumn{6}{c|}{\textbf{RoBERTa}} & \multicolumn{6}{c}{\textbf{ELECTRA}}\\
        \hline
        \textbf{Method} & \multicolumn{3}{c}{\textbf{\SDM}} & \multicolumn{3}{c|}{\textbf{DOMINO}} & \multicolumn{3}{c}{\textbf{\SDM}} &
        \multicolumn{3}{c|}{\textbf{DOMINO}} & \multicolumn{3}{c}{\textbf{\SDM}} &
        \multicolumn{3}{c}{\textbf{DOMINO}} \\
        \hline
        \textbf{Metric} & \textbf{C-prop} & \textbf{C-imp} & \textbf{imp} & \textbf{C-prop} & \textbf{C-imp} & \textbf{imp} & \textbf{C-prop} & \textbf{C-imp} & \textbf{imp} & \textbf{C-prop} &  \textbf{C-imp} & \textbf{imp} & \textbf{C-prop} & \textbf{C-imp} & \textbf{imp} & \textbf{C-prop} &  \textbf{C-imp} & \textbf{imp} \\
        \hline
        CoLA & \textbf{100.00} & \textbf{0.40} & \textbf{1.58} & 16.06 & -0.62 & 0.99 & \textbf{100.00} & \textbf{1.58} & \textbf{3.25} & 76.92 & -0.96 & -1.82 & \textbf{66.66} & \textbf{0.00} & \textbf{0.13} & 40.98 & -1.02 & 0.99\\
        QNLI & \textbf{100.00} & \textbf{0.45} & \textbf{2.07} & 3.92 & -0.42 & 0.56 & \textbf{100.00} & \textbf{0.65} & \textbf{1.88} & 34.70 & -0.30 & 1.15 & \textbf{97.80} & \textbf{0.19} & \textbf{0.86} & 30.10 & -0.26 & 0.52\\
        QQP & \textbf{99.83} & \textbf{0.19} & \textbf{1.25} & 0.85 & -0.35 & 1.16 & \textbf{49.85} & \textbf{-0.05} & 0.40 & 0.10 & -0.31 & \textbf{0.82} & \textbf{100.00} & \textbf{0.06} & \textbf{0.18} & 0.22 & -0.44 & 0.83 \\
        SST-2 & \textbf{96.30} & \textbf{0.35} & \textbf{0.86} & 71.43 & -0.24 & 0.66 & \textbf{91.30} & \textbf{0.00} & \textbf{0.87} & 9.09 & -0.60 & 0.75 & 69.44 & \textbf{0.24} & \textbf{1.22} & \textbf{75.86} & 0.12 & 0.90\\
        MNLI & 6.04 & \textbf{0.11} & \textbf{1.48} & \textbf{6.45} & -0.40 & 0.10 & \textbf{31.52} & \textbf{-0.06} & \textbf{1.14} & 0.92 & -1.46 & 1.08 & \textbf{97.27} & \textbf{0.00} & \textbf{0.49} & 1.60 & -0.32 & 0.22 \\
        SST-5 & \textbf{67.47} & \textbf{1.10} & \textbf{5.06} & 0.13 & -0.70 & 4.32 & \textbf{26.20} & \textbf{-1.94} & \textbf{4.77} & 0.25 & -1.97 & 3.90 & \textbf{100.00} & \textbf{0.38} & 0.96 & 2.71 & -0.57 & \textbf{3.40}\\
        Jigsaw-gender & 99.66 & 6.91 & 9.96  & \textbf{100.00} & \textbf{13.91} & \textbf{18.00} & 97.67 & \textbf{6.87} & \textbf{11.11} & \textbf{100.00} & 3.85 & 9.45 & 0.96 & -1.71 & \textbf{1.18} & \textbf{12.87} & \textbf{-0.58} & 0.66\\
        Jigsaw-race & \textbf{100.00} & \textbf{8.19} & \textbf{11.17} & 100.00 & 4.46 & 6.52 & \textbf{100.00} & \textbf{10.56} & \textbf{13.12} & 98.94 & 8.72 & 12.07 & \textbf{20.22} & \textbf{0.00} & \textbf{2.56} & 13.86 & -0.60 & 0.78\\
        Jigsaw-religion & \textbf{100.00} & \textbf{0.98} & \textbf{2.53} & 100.00 & 0.05 & 0.86 & 77.46 & 0.00 & \textbf{1.31} & \textbf{100.00} & \textbf{0.12} & 0.87 & \textbf{96.39} & \textbf{1.33} & \textbf{2.88} & 66.12 & 0.25 & 1.33\\
        \hline 
        \hline
        average & \textbf{85.48} & \textbf{2.08} & \textbf{4.00} & 44.32 & 1.74 & 3.69 & \textbf{74.89} & \textbf{1.96} & \textbf{4.21} & 46.75 & 0.85 & 3.14 & \textbf{72.08} & \textbf{0.05} & \textbf{1.16} & 27.15 & -0.38 & 1.07\\
        \bottomrule
    \end{tabular}
    }
    \vspace{-5pt}
    \caption{Selective Prediction Result}
    \label{tab:rejection}
    \vspace{-15pt}
\end{table*}

Each slice has one or more significant feature(s) as each datapoint may exhibit one or more error-prone feature(s). For each slice $S$, if $F$ is significant in $S$, $S$ is desirable to be as homogeneous as possible with regard to $F$ as we do not want to put datapoints with different features in one slice; $S$ is also desirable to be as complete as possible for $F$ as we want all error-prone datapoints featuring $F$ to be clustered in one slice. In addition, we also want to find as many error-correlated features as possible. Thus we propose four evaluation metrics: feature-prop which is the proportion of features in the benchmark that are detected to be significant for some slice, average homogeneity (Homo) which is the average Homo for each $F$ per slice featuring $F$, average completeness (Comp) which is the average Comp for each $F$ per slice featuring $F$, and average V-score\footnote{For each $F$ and each slice $S$ featuring $F$, a Homo score is defined by dividing $
|S_F|$ defined as $|\{d\in S\mid \mathcal{M}(d) \neq \text{label}(d)\text{ and } f(d) = 1\}|$ by $k$; a Comp score is defined by dividing $|S_F|$ by $|\{d\in D\mid \mathcal{M}(d) \neq \text{label}(d)\text{ and } f(d) = 1\}|$; a V-measure is computed as $\frac{2*\text{Homo}*\text{Comp}}{\text{Homo}+\text{Comp}}$.}.


We compare the performance using the average weighted (ave. weighted) V-score which is computed as follows:
\begin{align}
    \text{aver}&\text{age weighted V-score} = \nonumber \\
    &\frac{\Sigma_{dataset=\mathcal{D}} \text{feature-prop}_\mathcal{D} * \text{V-score}_\mathcal{D}}{ \text{number of datasets}}
\end{align}

We notice that \SDM performs the best on all metrics. \SDM-$Z$ also performs well in homogeneity scores but poorly at the completeness scores, which may be due to the model tends to cluster all sentences with similar semantic information together. \SDM-$E$ performs the worst on all metrics.

\subsection{Improve: Experiment Result}
In this section, we use selective prediction, flipping, and active learning tasks to evaluate an SDM performance externally. For all three tasks, we compare \SDM with confidence baseline as it is the second accurate baseline in finding error-prone datapoints shown in Table \ref{tab:selected_wrong_performance}.

\subsubsection{Selective Prediction Result}
We evaluate selective prediction performance by two metrics: \textit{proportion} and \textit{improvement}. \textit{Proportion} is the proportion of total step numbers where an SDM outperforms the baseline model in model accuracy. 
When the metric result is equal to 50.00$\%$, it means only half of the time SDM performs better than the baseline and thus the SDM is no better than the baseline; when it is higher than 50.00$\%$, then most of the time the SDM is more effective than the baseline model. \textit{Improvement} is the final efficacy improvement compared with the original efficacy. \textit{C-proportion} and \textit{C-improvement} are metrics that adopt confidence as the baseline model for comparison. 

For the confidence baseline, we reorder the datapoints by the confidence score from low to high and rejects the top-$\lvert\mathcal{E}_{SDM}\rvert$ datapoints.

\begin{table*}[t!]
    \centering
    \resizebox{17cm}{!}{
    \begin{tabular}{c|cccccc|cccccc|cccccc}
    \toprule
        \textbf{NLP model} & \multicolumn{6}{c|}{\textbf{BERT}} & \multicolumn{6}{c|}{\textbf{RoBERTa}} & \multicolumn{6}{c}{\textbf{ELECTRA}}\\
        \hline
        \textbf{Method} & \multicolumn{3}{c}{\textbf{\SDM}} & \multicolumn{3}{c|}{\textbf{DOMINO}} & \multicolumn{3}{c}{\textbf{\SDM}} &
        \multicolumn{3}{c|}{\textbf{DOMINO}} & \multicolumn{3}{c}{\textbf{\SDM}} &
        \multicolumn{3}{c}{\textbf{DOMINO}} \\
        \hline
        \textbf{Metric} & \textbf{C-prop} & \textbf{C-imp} & \textbf{imp} & \textbf{C-prop} & \textbf{C-imp} & \textbf{imp} & \textbf{C-prop} & \textbf{C-imp} & \textbf{imp} & \textbf{C-prop} &  \textbf{C-imp} & \textbf{imp} & \textbf{C-prop} & \textbf{C-imp} & \textbf{imp} & \textbf{C-prop} &  \textbf{C-imp} & \textbf{imp} \\
        \hline
        CoLA & \textbf{100.00} & \textbf{0.58} & \textbf{0.86} & 14.81 & -1.15 & -2.68 & \textbf{100.00} & \textbf{2.68} & \textbf{2.01} & 76.92 & -0.96 & -1.82 & \textbf{66.66} & \textbf{0.00} & \textbf{0.00} & 39.3 & -2.11 & -2.49 \\
        QNLI & \textbf{100.00} & \textbf{0.81} & \textbf{1.13} & 3.43 & -0.84 & -0.64 & \textbf{100.00} & \textbf{1.21} & \textbf{0.57} & 34.48 & -0.55 & -1.89 & \textbf{97.80} & \textbf{0.37} & \textbf{0.33} & 29.13 & -0.55 & -0.51 \\
        QQP & \textbf{99.83} & \textbf{0.36} & \textbf{0.12} & 0.81 & -0.66 & -0.89 & \textbf{49.85} & -0.09 & -0.06 & 0.94 & -0.60 & -0.79 & \textbf{100.00} & \textbf{0.12} & \textbf{0.07} & 0.15 & -0.84 & -0.95\\
        SST-2 & \textbf{96.30} & \textbf{0.46} & \textbf{0.00} & 68.57 & -0.46 & -1.15 & \textbf{91.30} & \textbf{0.00} & \textbf{0.00} & 60.61 & -1.15 & -0.69 & 69.44 & \textbf{0.46} & \textbf{-0.11} & \textbf{75.86} & 0.23 & -0.46\\
        MNLI & \textbf{52.64} & \textbf{0.99} & \textbf{0.04} & 52.90 & -0.50 & -0.79 & \textbf{31.52} & \textbf{-0.05} & \textbf{0.02} & 0.51 & -1.69 & -2.52 & \textbf{100.00} & \textbf{0.04} & 0.02 & 1.32 & -0.66 & \textbf{0.12}\\
        SST-5 & \textbf{77.47} & \textbf{2.67} & \textbf{0.27} & 62.44 & 0.98 & 0.02 & \textbf{70.00} & \textbf{0.00} & \textbf{0.18} & 68.18 & \textbf{0.00} & \textbf{0.08} & \textbf{68.90} & \textbf{0.18} & 0.13 & 2.38 & -0.74 & \textbf{0.27}\\
        Jigsaw-gender & 99.66 & 9.51 & 9.64 & \textbf{100.00} & \textbf{16.60} & \textbf{16.17} & 97.68 & \textbf{8.81} & \textbf{10.67} & \textbf{100.00} & 4.46 & 6.53 & 0.64 & -3.17 & -4.17 & \textbf{12.12} & -\textbf{1.13} & \textbf{-1.32}\\
        Jigsaw-race & \textbf{100.00} & 12.37 & 11.66 & \textbf{100.00} & \textbf{15.58} & \textbf{12.98} & \textbf{100.00} & \textbf{14.92} & \textbf{13.48} & 98.84 & 11.00 & 9.09 & \textbf{19.94} & -1.42 & \textbf{-1.02} & 13.25 & \textbf{-1.13} & -1.16 \\
        Jigsaw-religion & \textbf{100.00} & \textbf{1.87} & \textbf{1.45} & \textbf{100.00} & 0.30 & 0.18 & 77.46 & 0.00 & -0.57 & \textbf{100.00} & \textbf{0.18} & \textbf{-0.09} & \textbf{96.39} & \textbf{2.47} & \textbf{2.86} & 66.12 & 0.48 & 0.78\\ 
        \hline 
        \hline
        average & \textbf{91.77} & \textbf{3.29} & \textbf{2.80} & 55.87 & 3.28 & 2.58 & \textbf{78.87} & \textbf{4.71} & \textbf{2.92} & 60.06 & 1.19 & 0.21 & \textbf{68.86} & \textbf{-0.11} & \textbf{-0.21} & 26.63 & -0.72 & -0.64\\
        \bottomrule
    \end{tabular}
    }
    \vspace{-5pt}
    \caption{Flipping result}
    \label{tab:flipping}
    \vspace{-15pt}
\end{table*}

Table \ref{tab:rejection} reports the result on \SDM across the three models on all datasets: the average \textit{C-proportion} is 77.48 (higher than 50.00), \textit{C-improvement} is 1.36 and \textit{improvement} is 3.12, all demonstrating the advantage of \SDM.

\begin{figure}[t]
  \centering
    \begin{minipage}[h]{0.49\linewidth}
    \hspace{0.5cm}
    \includegraphics[width=1.5in]{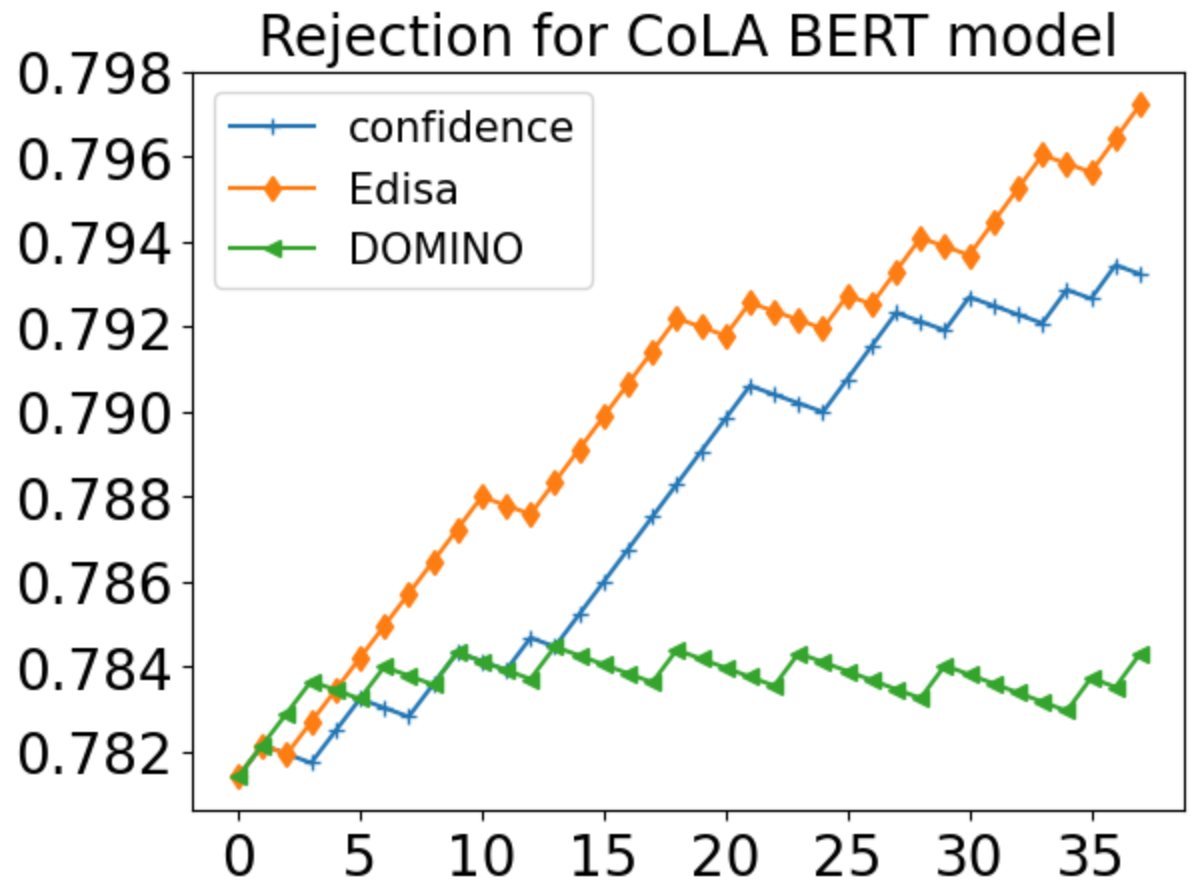}
    \includegraphics[width=1.5in]{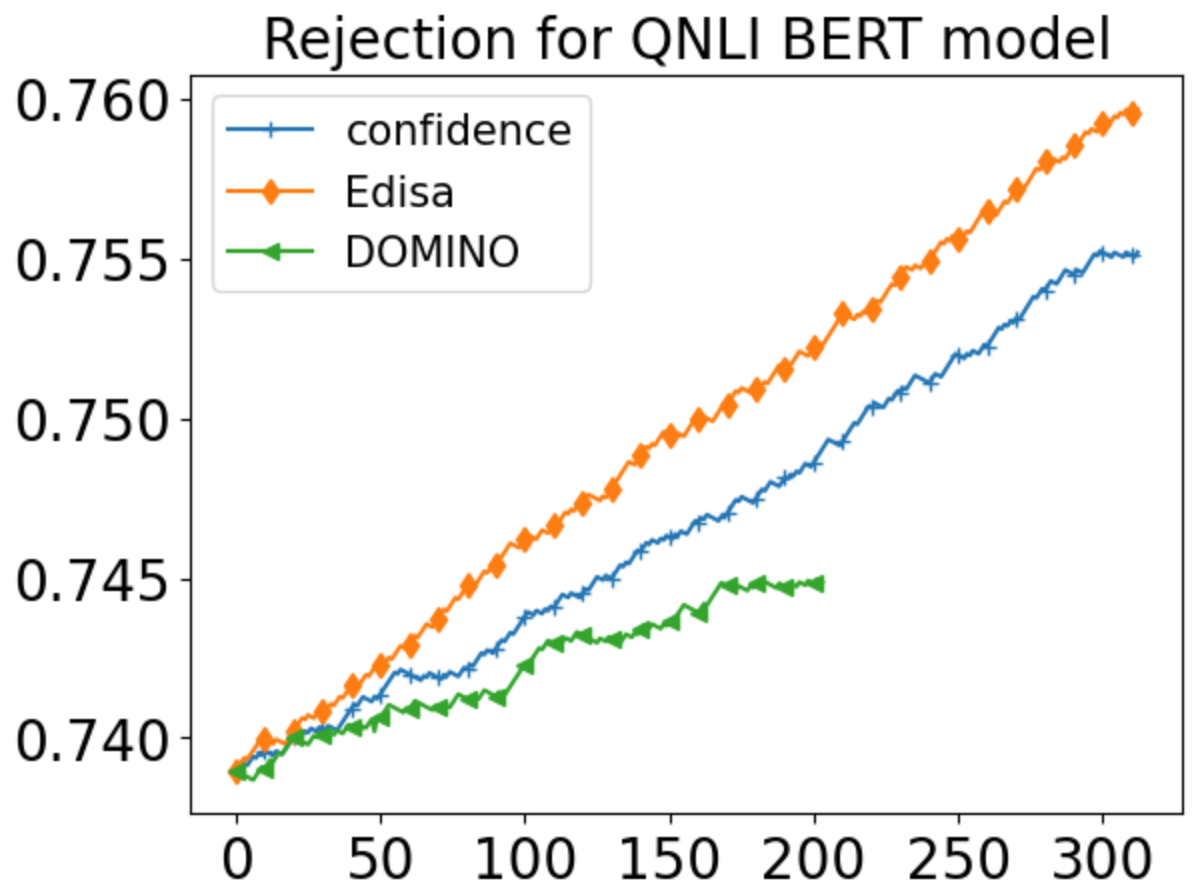}
    \end{minipage}
    \begin{minipage}[h]{0.49\linewidth}
    \hspace{1.5cm}
    \includegraphics[width=1.5in]{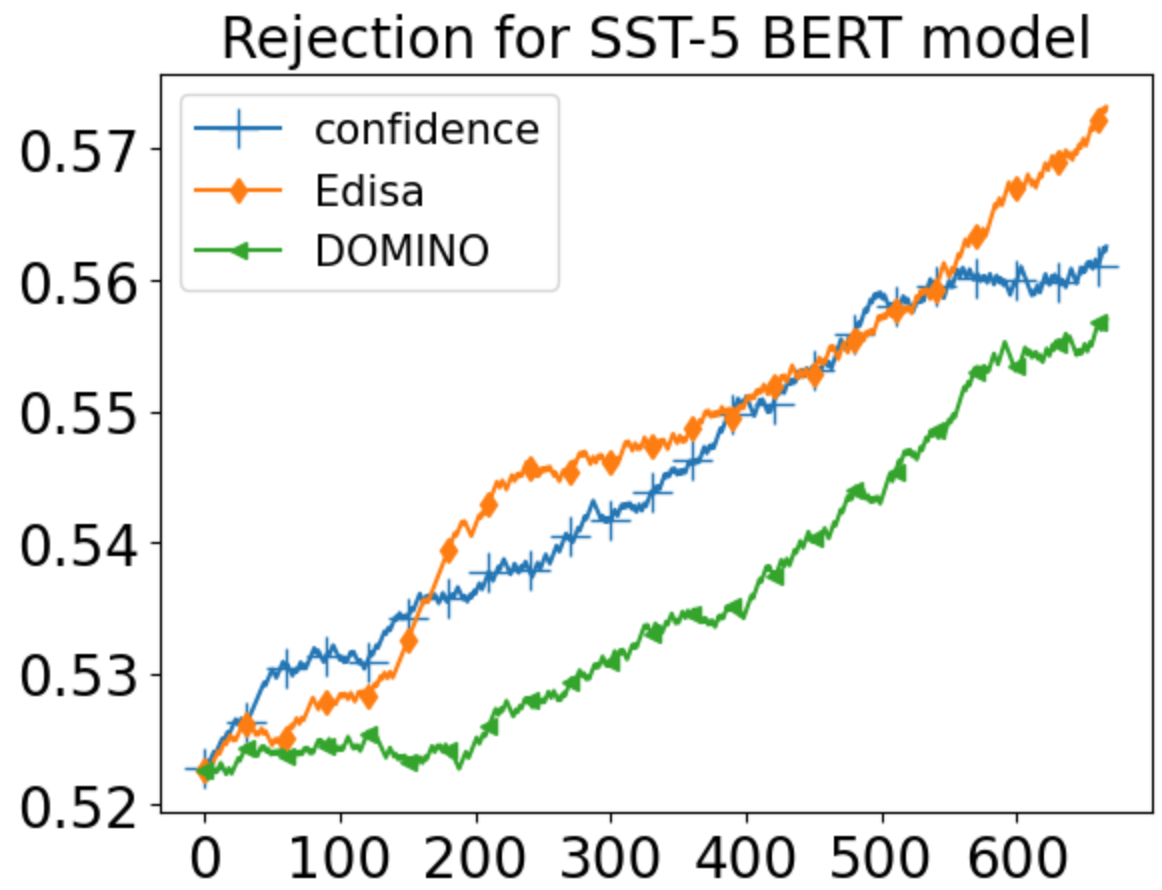}
    \includegraphics[width=1.5in]{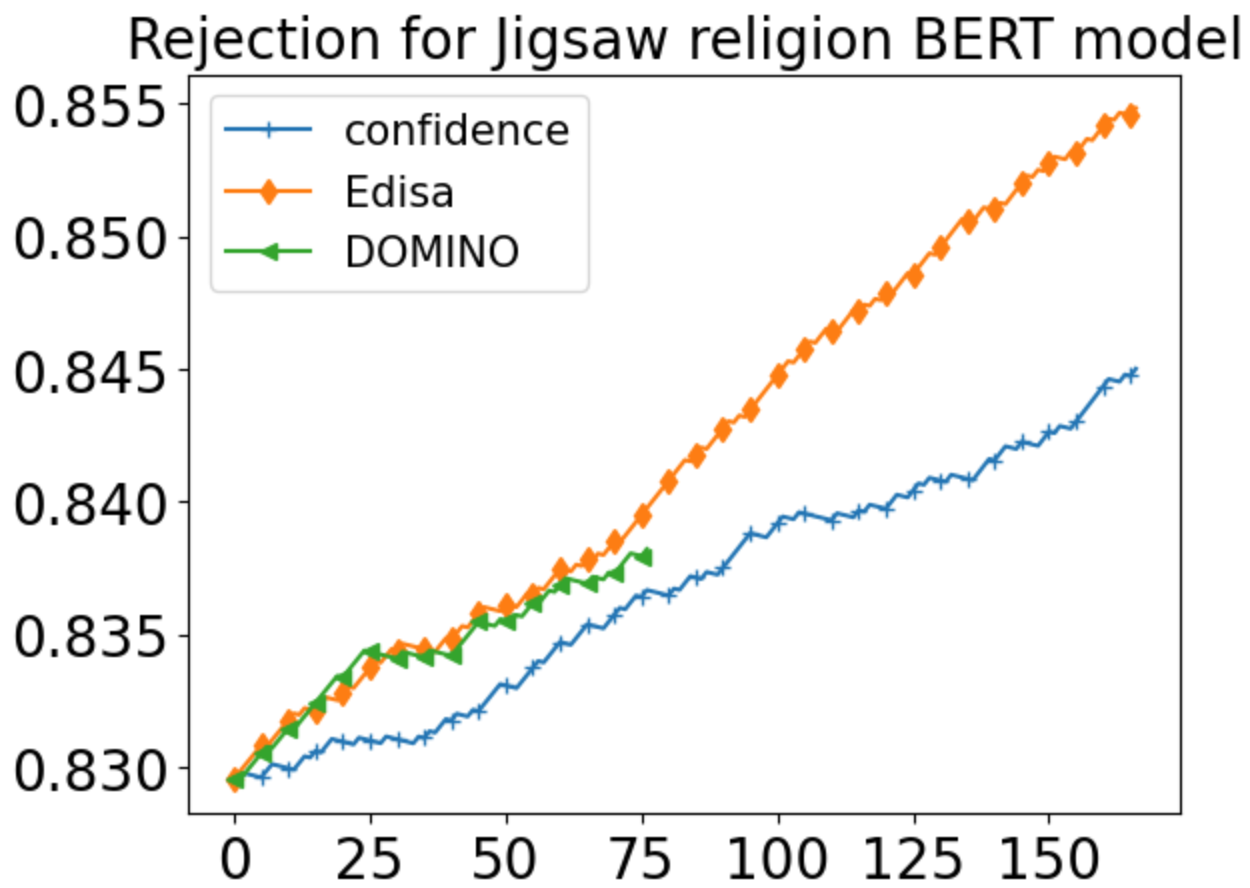}
    \end{minipage}
\vspace{-10pt}
\caption{Graphs for selective prediction task using confidence baseline \SDM model: CoLA, QNLI, SST-5, Jigsaw-religion. The x-axis is the number of rejected datapoints; the y-axis is the model efficacy.}
\label{fig:rejection}
\end{figure}

Visualization on the results of four diverse datasets based on BERT models are presented in Figure \ref{fig:rejection}: CoLA, QNLI, SST-5, and Jigsaw-religion, where CoLA and QNLI come from the GLUE benchmark; SST-5 is a multi-class dataset, and Jigsaw-religion comes from Jigsaw dataset. In each figure, the x-axis represents the number of datapoints rejected and the y-axis represents the efficacy of the remaining dataset. They demonstrate 
the change of efficacy stepwise comparing \SDM and confidence baseline: \SDM performs better at almost all steps, showing that it can always pick error-prone datapoints more accurately.

\subsubsection{Flipping Result}
The flipping task uses the same metrics as in the selective prediction task. 
Notice that SST-5 and MNLI are multi-class datasets: for SST-5, the validated confidence threshold to flip an error-prone datapoint is 0.35 for BERT, 0.37 for RoBERTa, and 0.5 for ELECTRA; for MNLI, the validated confidence threshold is 0.7 for BERT, 0.35 for RoBERTa, and 0.5 for ELECTRA. Based on Table \ref{tab:flipping}, the average \textit{C-proportion} is 79.83 (above 50.00), average \textit{C-improvement} is 2.63 and average \textit{improvement} is 1.84, showing that \SDM is able to improve the model directly.

\begin{figure}[t]
  \centering
    \begin{minipage}[h]{0.49\linewidth}
    \hspace{0.5cm}
    \includegraphics[width=1.5in]{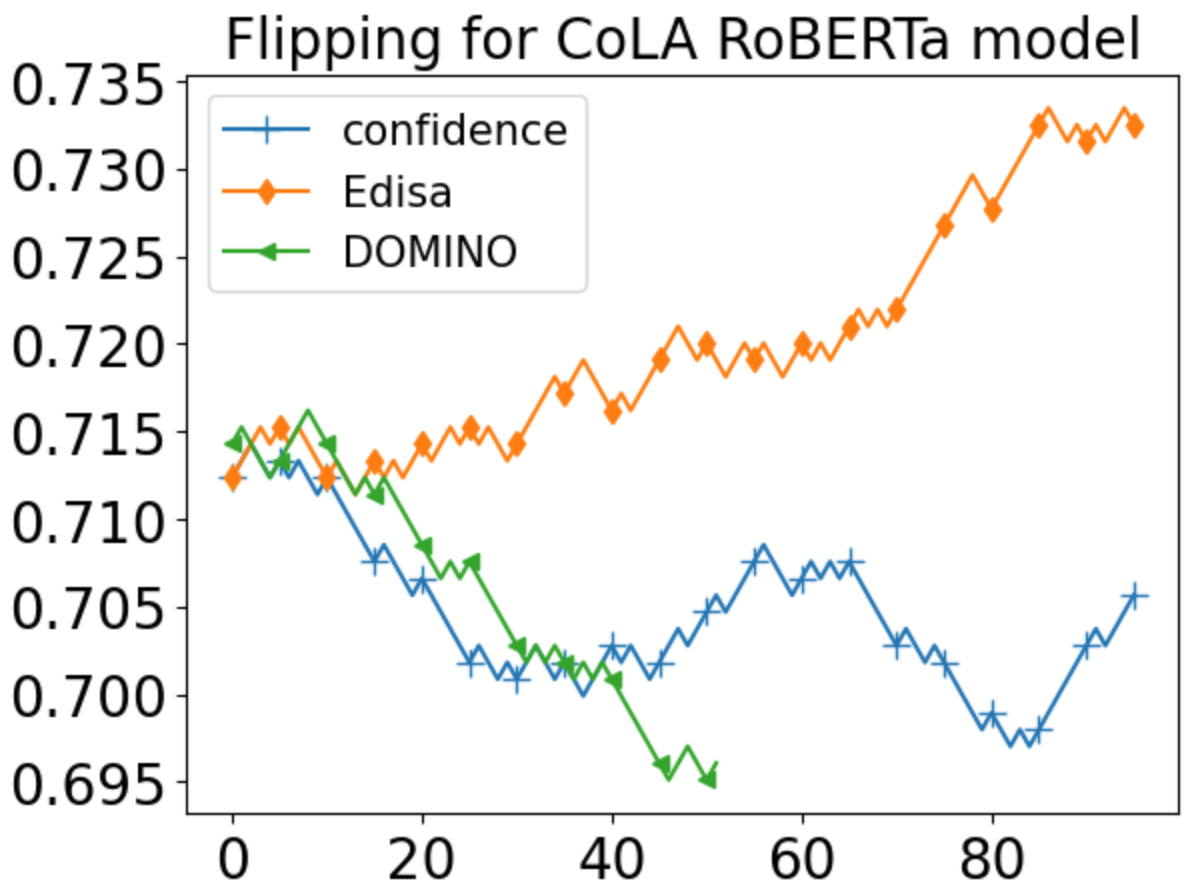}
    \includegraphics[width=1.5in]{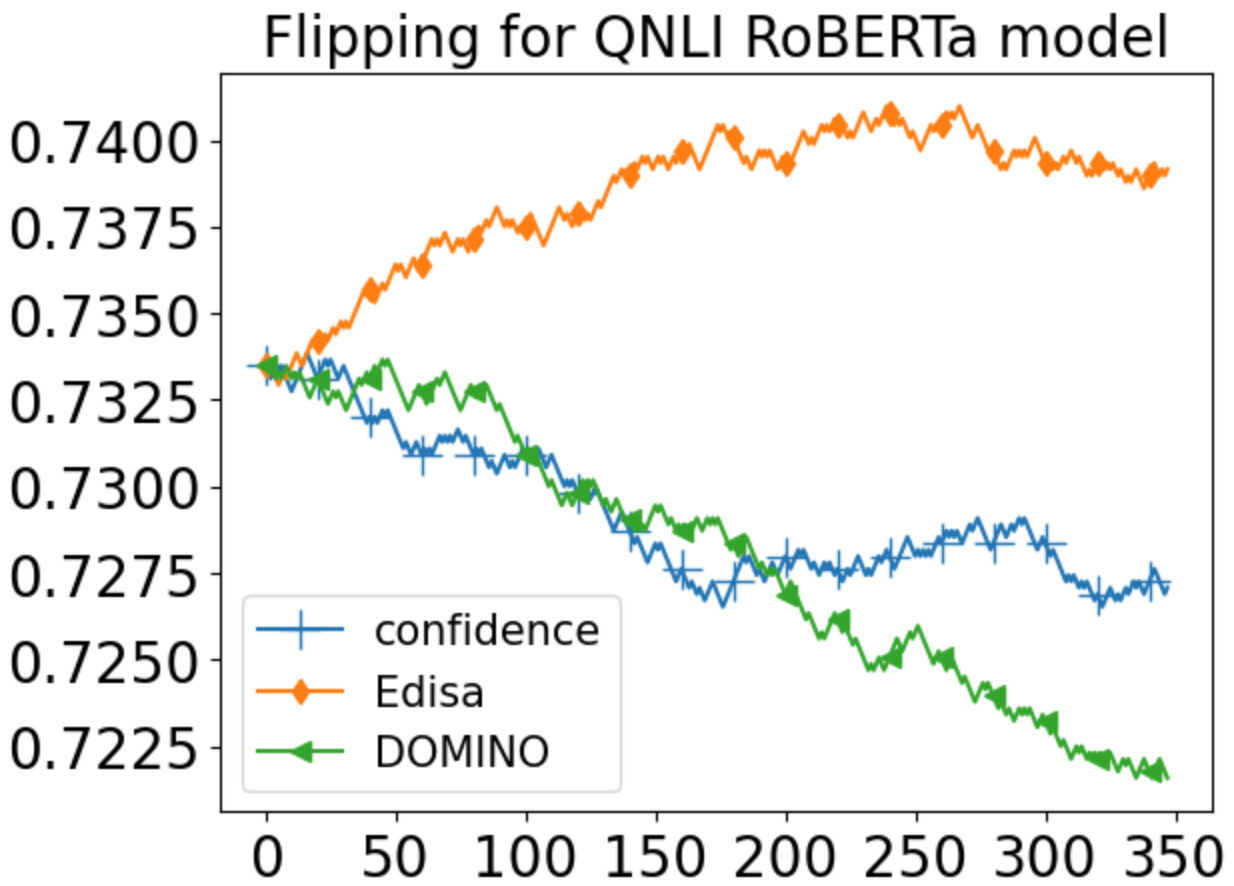}
    \end{minipage}
    \begin{minipage}[h]{0.49\linewidth}
    \hspace{1.5cm}
    \includegraphics[width=1.5in]{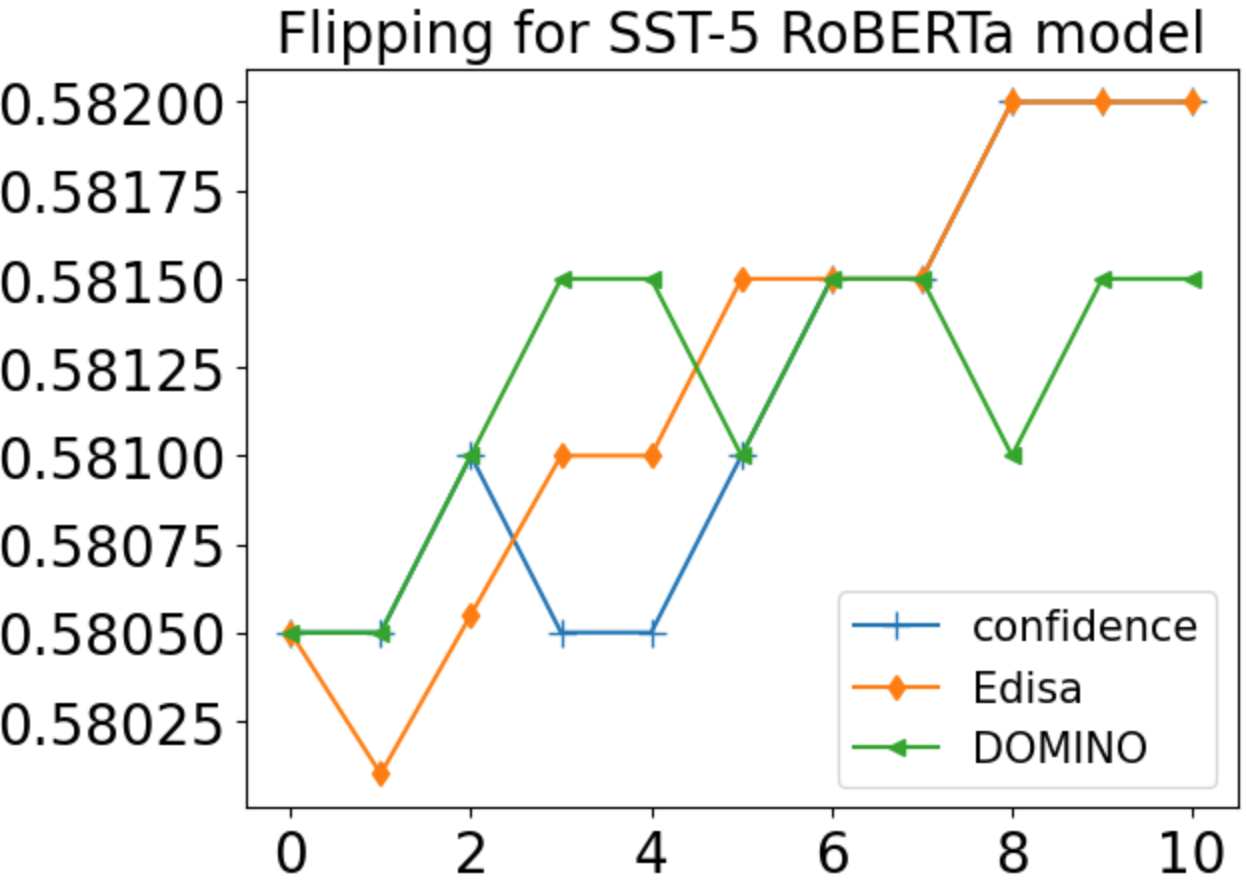}
    \includegraphics[width=1.5in]{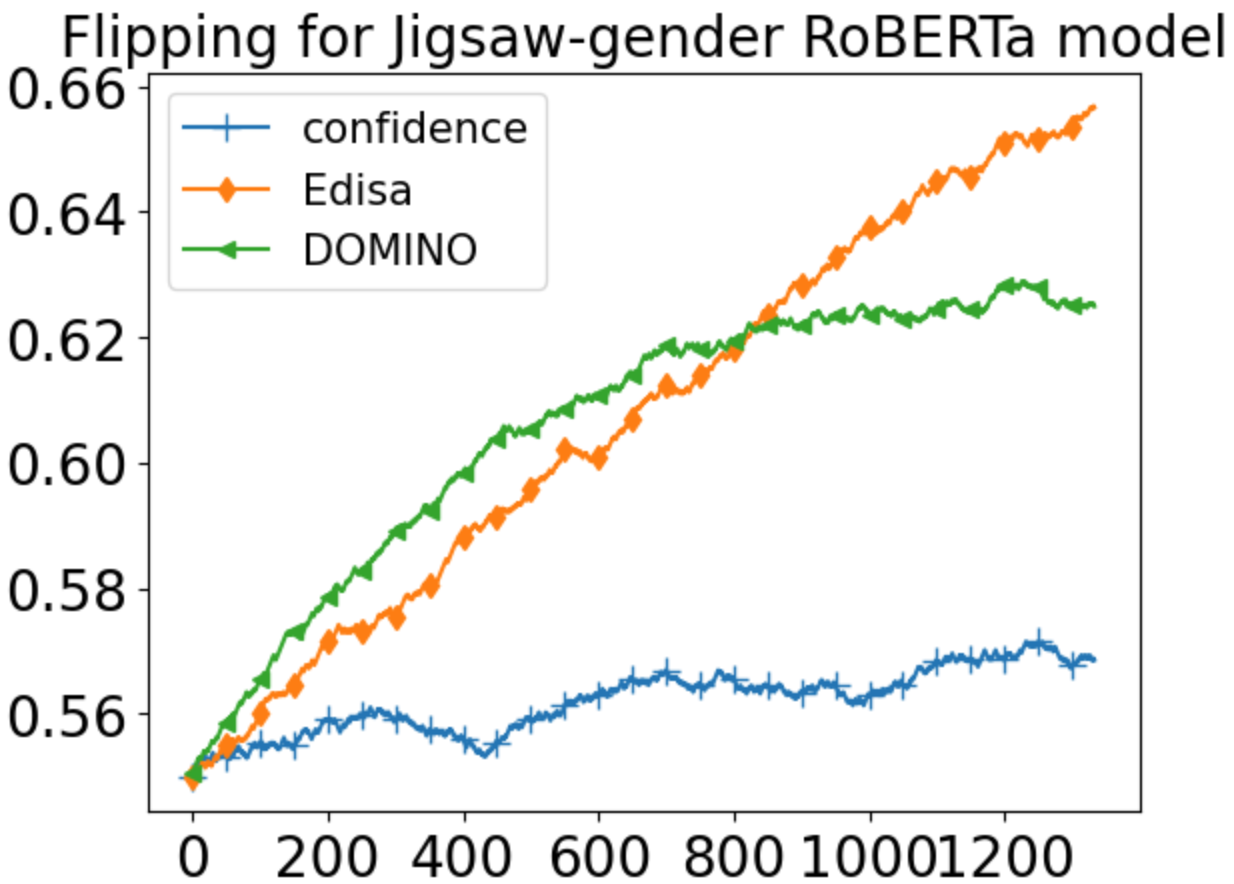}
    \end{minipage}
\vspace{-10pt}
\caption{Graphs for the flipping task using confidence baseline and \SDM model: CoLA, QNLI, SST-5, Jigsaw-gender. The x-axis is the number of flipped datapoints; the y-axis is the model efficacy.}
\label{fig:flipping}
\end{figure}

Figure \ref{fig:flipping} contains four graphs of flipping on RoBERTa models: \SDM performs better on almost all steps and can indeed help model performance on the original test dataset. On the contrary, the confidence baseline and DOMINO baseline are not efficacious enough in selecting error-prone datapoints to improve model performance: the efficacy performance either holds almost constant or decreases.

\subsubsection{Active Learning Result}
In active learning simulation, we adopt three other simulations as baselines: DOMINO, confidence learning, and random learning. Confidence learning selects a certain number of low-confidence extra datapoints to train per step; random learning randomly selects a certain number of extra datapoints to train per step.

\begin{figure}[ht]
    \centering
    \includegraphics[scale=0.3]{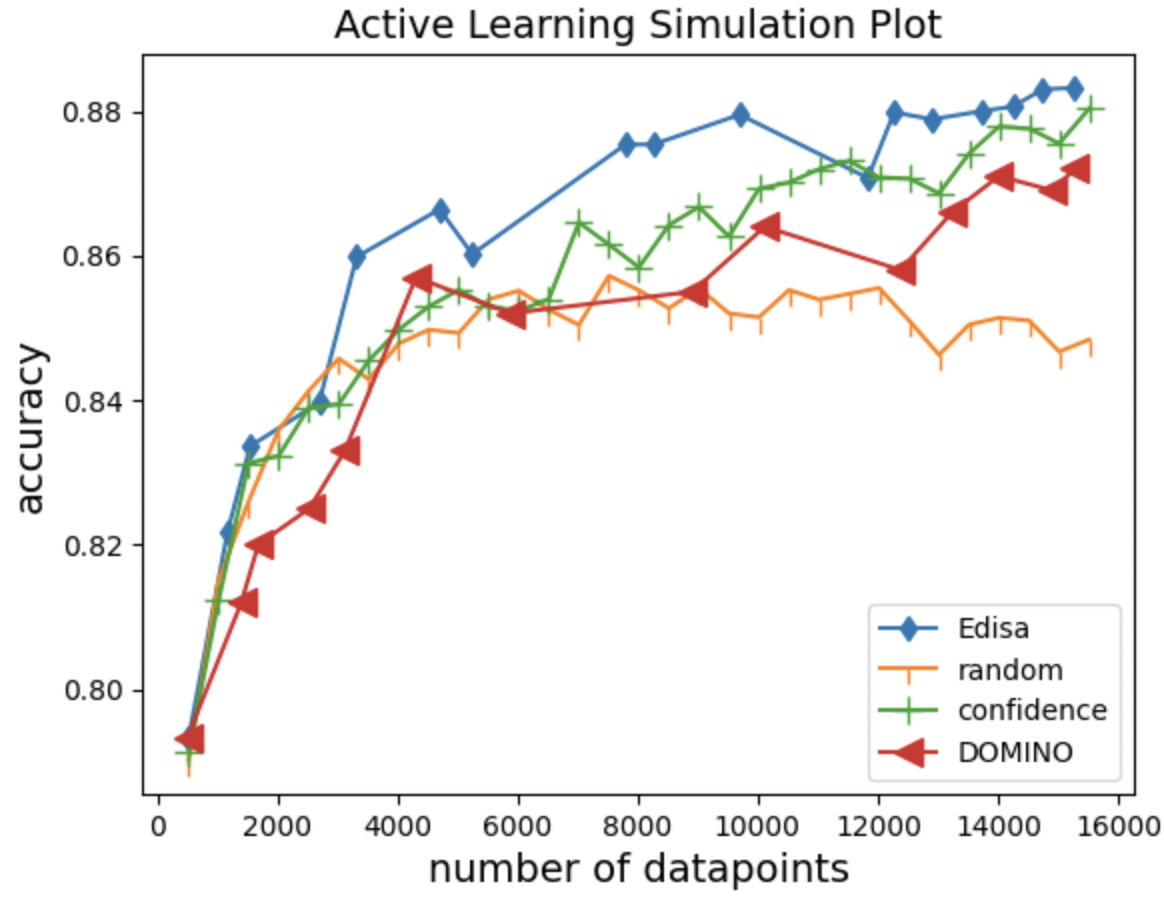}
    \vspace{-10pt}
    \caption{Active learning on QNLI dataset}
    \label{fig:active_learning}
\end{figure}

We demonstrate performance on this task by working on the QNLI BERT model in Figure.\ref{fig:active_learning}. The x-axis is the number of datapoints used to train and the y-axis is the NLP model's accuracy. We use 1$\%$ datapoints of the original training dataset as seed. For confidence learning and random learning, we select 500 more datapoints for each step; for active learnings with \SDM and DOMINO, the SDM (\SDM or DOMINO) decides how many extra datapoints to train on. All active learning processes have run 10 times with different random seeds using up to 16k datapoints (about 30 learning steps) when active learning and confidence learning converge. The y-axis demonstrates the average accuracies in the 10 experiments.

The figure demonstrates that active learning using \SDM performs noticeably better. \SDM and confidence learning converge to similar accuracy after learning on 16k datapoints. We perform the paired Student's t-test protocol with $p$-value $<0.05$ to show that the \SDM process's accuracies on the steps from 3k to 10k datapoints are significantly higher than accuracies of baselines.

\section{Conclusion}
In this paper, we take the first step to build a comprehensive slice detection framework \name on NLP with principled evaluation tasks, linguistic tools, and metrics. It discovers error-prone datapoints, clusters datapoints in an error slice under the interpretable concept, and directly improves model performance on unlabeled datasets. It shows that discovering error slices can provide not only insights into model behaviors but also actionable and automatic model improvement methods. Experiments show that \SDM is a more efficacious model than current baselines. We hope this benchmark can facilitate further research in SDM.



\section{Limitation and Future Work}
This study presents an all-encompassing benchmark designed to evaluate slice detection models from three distinct perspectives. As a pioneering endeavor in benchmarking SDMs, it is important to recognize certain limitations, which provide avenues for improvement in subsequent research:
\begin{enumerate}
    \item The \SDM model currently only works for encoder-only models, while not directly applicable to encoder-decoder models such as T5 and the prevalent decoder-only models such as GPT series models. Future work should extend slice detection models such as \SDM to more model structures.
    \item The \SDM model currently focuses on classification datasets. Future work should consider extending to tasks such as logistic regression and text generation.
    \item The \SDM model assigns each datapoint to one slice and \name benchmark assumes that a single feature can represent each slice. This oversimplification may not suffice for the intricacies of expansive language models prevalent in today's NLP landscape. Future endeavors should contemplate refining this approach in the SDM and evaluation by either (1) attributing each data point to multiple slices, or (2) denoting each slice with several features, or a combination of both.
\end{enumerate}

We hope forthcoming research can be built based on \SDM model and the benchmark and thereby deepening our understanding of model performance.

\section*{Acknowledgements}
I extend my gratitude to my co-authors for their invaluable contributions, including discussions and experimental suggestions. I also appreciate the thoughtful feedback from the reviewers and the guidance of the editor, which greatly enhanced the quality of this manuscript.

\bibliographystyle{acl_natbib}
\bibliography{custom}

\clearpage

\appendix

\section{Appendix}

\subsection{Datasets in Experiments}
\textbf{CoLA}: The Corpus of Linguistic Acceptability is a binary classification dataset aiming at distinguishing ungrammatical sentences from grammatical sentences, consisting of 10657 sentences from 23 linguistics publications.\\
\textbf{QNLI}: The Question-answering Natural Language Inference dataset is a binary classification dataset aiming at judging whether the context sentence contains the answer to the question, automatically derived from the Stanford Question Answering Dataset v1.1. \\
\textbf{QQP}: Quora Question Pairs dataset is a binary classification task aiming at judging whether the two questions are paraphrases of each other, consisting of over 400,000 question pairs. \\
\textbf{SST-2}: The Stanford Sentiment Treebank is a binary classification task analyzing the effects of sentiment consisting of 215,154 sentences.\\
\textbf{MNLI}: The Multi-Genre Natural Language Inference corpus is a thee-class classification task consisting of 433k sentence pairs annotated with textual entailment information.\\
\textbf{SST-5}: The Stanford Sentiment Treebank is a fine-grained five-class classification task analyzing the effects of sentiment in language.

The above datasets are based on GLUE. We did not train on other GLUE datasets due to their small training data size, such as RTE and WNLI.

\textbf{Jigsaw}: The Jigsaw dataset is a binary classification dataset aiming to detect toxic comments and minimize unintended model bias consisting of about 180k datapoints. We constructed three sub-datasets based on Jigsaw, each focusing on one type of potential model bias: gender (male, female and other$\_$gender), race (black, white, Asian, etc.), and religion (atheist, Christian, Muslim, etc.). We also re-balance the dataset so that 50$\%$ of the datapoints have a non-trivial value on at least one feature. The \textbf{Jigsaw-gender} consists of 37k datapoints, \textbf{Jigsaw-race} consists of 40k datapoints, \textbf{Jigsaw-religion} consists of 183k datapoints.

\if and only ifalse
\subsection{Extra Experiment Results}
\begin{table*}[t!]
    \centering
    \resizebox{18cm}{!}{
    \begin{tabular}{c|c|ccc|ccc|ccc|ccc|ccc|ccc|ccc|ccc}
    \toprule
        \textbf{dataset} & \textbf{model} & \multicolumn{3}{c|}{\textbf{length}} & \multicolumn{3}{c|}{\textbf{negation}} & \multicolumn{3}{c|}{\textbf{reflexive}} & \multicolumn{3}{c|}{\textbf{how-Q}} & \multicolumn{3}{c|}{\textbf{why-Q}} & \multicolumn{3}{c|}{\textbf{MP}} & \multicolumn{3}{c|}{\textbf{NS}} & \multicolumn{3}{c|}{\textbf{echo-Q}} \\
        \hline
        \textbf{Metric} & & \textbf{V-score} & \textbf{homo} & \textbf{comp} & \textbf{V-score} & \textbf{homo} & \textbf{comp} & \textbf{V-score} & \textbf{homo} & \textbf{comp} & \textbf{V-score} & \textbf{homo} & \textbf{comp} & \textbf{V-score} & \textbf{homo} & \textbf{comp} & \textbf{V-score} & \textbf{homo} & \textbf{comp} & \textbf{V-score} & \textbf{homo} & \textbf{comp} & \textbf{V-score} & \textbf{homo} & \textbf{comp} \\
        \hline
        CoLA & \SDM & 6.78 & 3.52 & 92.59 & 38.09 & 25.00 & 80.00 & 8.78 & 4.61 & 90.00 & 4.40 & 2.30 & 50.00 & N/A & N/A & N/A & 35.10 & 21.95 & 87.50 & 17.27 & 10.00 & 63.16 & 57.14 & 40.00 & 100.00\\
        \hdashline
        CoLA & DOMINO & 8.83 & 4.64 & 91.67 & 9.91 & 5.22 & 96.30 & 8.90 & 4.71 & 81.81 & N/A & N/A & N/A & N/A & N/A & N/A & 12.98 & 7.03 & 84.85 & 22.43 & 13.04 & 80.00 & N/A & N/A & N/A\\
        \hdashline
        CoLA & \SDM-$Z$ &6.93 & 3.60 & 93.75 & N/A & N/A & N/A & N/A & N/A & N/A & N/A & N/A & N/A & N/A & N/A & N/A & 15.67 & 8.67 & 81.82 & 40.00 & 37.50 & 42.86 & N/A & N/A & N/A \\
        \hdashline
        CoLA & \SDM-$E$ &5.62 & 2.93 & 70.37 & N/A & N/A & N/A & 8.63 & 4.55 & 84.21 & 2.41 & 1.22 & 100.00 & N/A & N/A & N/A & 17.39 & 10.53 & 50.00 & 15.46 & 9.68 & 38.46 & N/A & N/A & N/A \\
        \hline
        QNLI & \SDM & 7.40 & 3.95 & 58.16 & 1.10 & 0.56 & 100.00 & N/A & N/A & N/A & 5.79 & 3.10 & 43.48 & 1.63 & 0.83 & 63.27 & 7.53 & 4.03 & 58.21 & N/A & N/A & N/A & N/A & N/A & N/A \\
        \hdashline
        QNLI & DOMINO & 8.38 & 4.47 & 66.01 & 3.26 & 1.68 & 61.54 & 0.31 & 0.15 & 81.82 & 8.20 & 4.40 & 59.31 & 2.17 & 1.10 & 89.47 & 8.00 & 4.30 & 56.71 & N/A & N/A & N/A & 0.38 & 0.19 & 100.00\\
        \hdashline
        QNLI & \SDM-$Z$ &N/A & N/A & N/A & 0.38 & 0.19 & 70.00 & N/A & N/A & N/A & N/A & N/A & N/A & N/A & N/A & N/A & N/A & N/A & N/A & N/A & N/A & N/A & N/A & N/A & N/A\\
        \hdashline
        QNLI & \SDM-$E$ &7.11 & 3.79 & 57.42 & 3.41 & 1.75 & 61.54 & N/A & N/A & N/A & 9.81 & 5.66 & 36.67 & 1.65 & 0.84 & 56.14 & 15.11 & 8.42 & 73.33 & N/A & N/A & N/A & N/A & N/A & N/A \\
        \hline
        QQP & \SDM & 10.79 & 5.83 & 72.00 & 3.32 & 1.69 & 78.57 & 4.17 & 2.13 & 100.00 & 15.37 & 9.01 & 52.27 & 7.14 & 3.85 & 50.00 & 1.67 & 0.85 & 47.62 & N/A & N/A & N/A & 1.40 & 0.71 & 68.75\\
        \hdashline
        QQP & DOMINO & 3.90 & 2.03 & 50.93 & 41.80 & 27.28 & 89.47 &  N/A & N/A & N/A & 7.74 & 4.13 & 61.39 & 26.36 & 15.52 & 87.50 & 12.76 & 7.02 & 70.00 & N/A & N/A & N/A & N/A & N/A & N/A\\
        \hdashline
        QQP & \SDM-$Z$ &N/A & N/A & N/A & N/A & N/A & N/A & N/A & N/A & N/A & 15.32 & 8.97 & 52.38 & N/A & N/A & N/A & N/A & N/A & N/A & N/A & N/A & N/A & N/A & N/A & N/A \\
        \hdashline
        QQP & \SDM-$E$ &15.44 & 8.70 & 100.00 & N/A & N/A & N/A & 3.28 & 1.67 & 100.00 & 18.63 & 10.75 & 69.70 & 10.60 & 5.69 & 77.14 & N/A & N/A & N/A &  N/A & N/A & N/A &  N/A & N/A & N/A \\
        \hline
        SST-2 & \SDM & 5.08 & 2.64 & 67.80 & 2.78 & 1.42 & 71.82 & N/A & N/A & N/A & N/A & N/A & N/A & N/A & N/A & N/A & 4.51 & 2.35 & 54.55 & 2.81 & 1.44 & 57.89 & N/A & N/A & N/A\\
        \hdashline
        SST-2 & DOMINO & 3.83 & 1.96 & 78.01 & 4.66 & 2.40 & 74.64 & N/A & N/A & N/A & N/A & N/A & N/A & N/A & N/A & N/A & 1.94 & 0.98 & 57.02 & 4.92 & 2.55 & 72.11 & N/A & N/A & N/A\\
        \hdashline
        SST-2 & \SDM-$Z$ &3.27 & 1.67 & 90.00 & 3.68 & 1.89 & 70.59 & N/A & N/A & N/A & N/A & N/A & N/A & N/A & N/A & N/A & N/A & N/A & N/A & 12.24 & 6.67 & 75.00 & N/A & N/A & N/A \\
        \hdashline
        SST-2 & \SDM-$E$ &5.13 & 2.74 & 40.00 & 2.36 & 1.20 & 64.18 & N/A & N/A & N/A & N/A & N/A & N/A & N/A & N/A & N/A & 1.47 & 0.75 & 52.94 & N/A & N/A & N/A & N/A & N/A & N/A \\
        \hline
        MNLI & \SDM & 12.53 & 6.87 & 71.43 & 19.70 & 11.59 & 65.64 & 2.22 & 1.14 & 50.00 & N/A & N/A & N/A & N/A & N/A & N/A & 7.02 & 3.83 & 42.11 & 7.11 & 3.85 & 47.06 & N/A & N/A & N/A \\
        \hdashline
        MNLI & DOMINO & 9.92 & 5.55 & 54.55 & 24.14 & 14.35 & 76.12 & 3.14 & 1.61 & 60.00 & N/A & N/A & N/A & N/A & N/A & N/A & N/A & N/A & N/A & 13.70 & 7.59 & 70.97 & N/A & N/A & N/A \\
        \hdashline
        MNLI & \SDM-$Z$ &9.62 & 5.26 & 55.56 & N/A & N/A & N/A & 24.40 & 15.00 & 65.42 & N/A & N/A & N/A & N/A & N/A & N/A & 11.57 & 6.54 & 50.00 & 15.38 & 14.29 & 16.67 & N/A & N/A & N/A\\
        \hdashline
        MNLI & \SDM-$E$ &6.98 & 3.73 & 54.43 & 3.15 & 1.60 & 100.00 & 22.70 & 13.59 & 68.88 & N/A & N/A & N/A & N/A & N/A & N/A & N/A & N/A & N/A & 16.74 & 9.46 & 72.73 & N/A & N/A & N/A \\
        \hline
        SST-5 & \SDM & 22.34 & 12.75 & 90.23 & 25.97 & 16.91 & 55.88 & 8.86 & 4.71 & 75.00 & N/A & N/A & N/A & N/A & N/A & N/A & 13.12 & 7.50 & 52.41 & 26.58 & 16.45 & 69.20 & 8.00 & 4.17 & 100.00 \\
        \hdashline
        SST-5 & DOMINO & 22.34 & 12.75 & 90.23 & 25.97 & 16.91 & 55.88 & 8.8.6 & 4.71 & 75.00 & N/A & N/A & N/A & N/A & N/A & N/A & 13.12 & 7.50 & 52.41 & 26.58 & 16.45 & 69.20 & 8.00 & 4.17 & 100.00 \\
        \hdashline
        SST-5 & \SDM-$Z$ &N/A & N/A & N/A & 12.50 & 50.00 & 7.14 & N/A & N/A & N/A & N/A & N/A & N/A & N/A & N/A & N/A & N/A & N/A & N/A & 18.18 & 22.22 & 15.38 & N/A & N/A & N/A\\
        \hdashline
        SST-5 & \SDM-$E$ &14.90 & 8.02 & 57.95 & 26.67 & 30.77 & 100.00 & 10.13 & 5.48 & 66.67 & 16.67 & 9.09 & 100.00 & N/A & N/A & N/A & N/A & N/A & N/A & 20.23 & 12.84 & 47.62 & N/A & N/A & N/A \\
        \hline
        \hline
        \textbf{dataset} & \textbf{model} & \multicolumn{3}{c|}{\textbf{tree-depth}} & \multicolumn{3}{c|}{\textbf{extra-to}} & \multicolumn{3}{c|}{\textbf{multi-modal}} & \multicolumn{3}{c|}{\textbf{quantifier}} & \multicolumn{3}{c|}{\textbf{voice}} & \multicolumn{3}{c|}{\textbf{tense}} & \multicolumn{3}{c|}{\textbf{comparison}} & \multicolumn{3}{c|}{\textbf{long-distance}} \\
        \hline
        \textbf{Metric} & & \textbf{V-score} & \textbf{homo} & \textbf{comp} & \textbf{V-score} & \textbf{homo} & \textbf{comp} & \textbf{V-score} & \textbf{homo} & \textbf{comp} & \textbf{V-score} & \textbf{homo} & \textbf{comp} & \textbf{V-score} & \textbf{homo} & \textbf{comp} & \textbf{V-score} & \textbf{homo} & \textbf{comp} & \textbf{V-score} & \textbf{homo} & \textbf{comp} & \textbf{V-score} & \textbf{homo} & \textbf{comp} \\
        \hline
        CoLA & \SDM & 32.43 & 22.22 & 60.00 & N/A & N/A & N/A & 7.27 & 3.77 & 100.00 & 9.68 & 5.12 & 85.71 & 15.38 & 8.33 & 100.00 & N/A & N/A & N/A & 14.89 & 8.11 & 90.91 & 18.75 & 12.50 & 37.50\\
        \hdashline
        CoLA & DOMINO & 20.20 & 11.86 & 68.00 & N/A & N/A & N/A & 44.44 & 28.57 & 100.00 & 11.46 & 6.12 & 90.32 & 11.59 & 6.56 & 50.00 & N/A & N/A & N/A & 3.27 & 1.67 & 86.36 & 19.83 & 11.59 & 68.42\\
        \hdashline
        CoLA & \SDM-$Z$ &43.48 & 33.33 & 62.50 & N/A & N/A & N/A & 46.15 & 30.00 & 100.00 & 14.29 & 7.69 & 100.00 & 1.83 & 0.93 & 87.50 & N/A & N/A & N/A & 25.00 & 25.00 & 25.00 & 27.78 & 20.83 & 41.67 \\
        \hdashline
        CoLA & \SDM-$E$ &8.00 & 4.35 & 50.00 & N/A & N/A & N/A & 7.89 & 4.11 & 100.00 & 12.22 & 6.57 & 87.88 & N/A & N/A & N/A & N/A & N/A & N/A & N/A & N/A & N/A & 7.74 & 4.35 & 35.29\\
        \hline
        QNLI & \SDM & N/A & N/A & N/A & N/A & N/A & N/A & N/A & N/A & N/A & N/A & N/A & N/A & 9.61 & 5.24 & 58.11 & N/A & N/A & N/A & 3.42 & 1.76 & 55.56 & 9.82 & 5.41 & 52.71 \\
        \hdashline
        QNLI & DOMINO & 4.67 & 2.43 & 61.29 & N/A & N/A & N/A & N/A & N/A & N/A & 8.36 & 4.51 & 57.94 & 8.65 & 4.65 & 61.23 & N/A & N/A & N/A & 4.92 & 2.57 & 56.94 & 7.77 & 4.13 & 52.53 \\
        \hdashline
        QNLI & \SDM-$Z$ &N/A & N/A & N/A & N/A & N/A & N/A & N/A & N/A & N/A & N/A & N/A & N/A & 10.32 & 5.73 & 52.00 & N/A & N/A & N/A & N/A & N/A & N/A & N/A & N/A & N/A \\
        \hdashline
        QNLI & \SDM-$E$ &N/A & N/A & N/A & N/A & N/A & N/A & N/A & N/A & N/A & N/A & N/A & N/A & 6.98 & 3.70 & 61.07 & N/A & N/A & N/A & 4.30 & 2.23 & 62.50 & 7.57 & 4.09 & 51.19 \\
        \hline
        QQP & \SDM & N/A & N/A & N/A & N/A & N/A & N/A & N/A & N/A & N/A & N/A & N/A & N/A & N/A & N/A & N/A & N/A & N/A & N/A &11.90 & 6.56 & 64.10 & N/A & N/A & N/A \\
        \hdashline
        QQP & DOMINO & N/A & N/A & N/A & N/A & N/A & N/A & N/A & N/A & N/A & 21.43 & 14.28 & 42.86 & 13.33 & 7.14 & 100.00 & N/A & N/A & N/A & 10.27 & 5.56 & 68.04 & 12.06 & 6.51 & 81.82  \\
        \hdashline
        QQP & \SDM-$Z$ &N/A & N/A & N/A & N/A & N/A & N/A & N/A & N/A & N/A & 9.60 & 5.41 & 42.86 & N/A & N/A & N/A & N/A & N/A & N/A & N/A & N/A & N/A & 21.28 & 13.51 & 50.00 \\
        \hdashline
        QQP & \SDM-$E$ &N/A & N/A & N/A & N/A & N/A & N/A & N/A & N/A & N/A & 14.66 & 8.67 & 46.67 & 4.43 & 2.27 & 85.71  & N/A & N/A & N/A & 10.53 & 5.70 & 69.39 & 7.86 & 4.20 & 60.61 \\
        \hline
        SST-2 & \SDM & 5.19 & 2.71 & 57.50 & N/A & N/A & N/A & N/A & N/A & N/A & 6.45 & 3.45 & 50.00 & N/A & N/A & N/A & N/A & N/A & N/A & 7.12 & 3.84 & 51.22 & 1.40 & 0.71 & 46.00\\
        \hdashline
        SST-2 & DOMINO & 5.19 & 2.71 & 57.50 & N/A & N/A & N/A & N/A & N/A & N/A & 5.70 & 2.97 & 70.67 & 2.36 & 1.20 & 85.71 & 3.03 & 1.55 & 65.31 & 5.29 & 2.74 & 74.44 & 1.31 & 0.66 & 59.14\\
        \hdashline
        SST-2 & \SDM-$Z$ &N/A & N/A & N/A & N/A & N/A & N/A  & N/A & N/A & N/A & N/A & N/A & N/A & 13.33 & 7.14 & 100.00 & N/A & N/A & N/A  & 5.92 & 3.13 & 57.14 & N/A & N/A & N/A  \\
        \hdashline
        SST-2 & \SDM-$E$ &7.14 & 4.17 & 25.00 & N/A & N/A & N/A & N/A & N/A & N/A & 2.90 & 1.49 & 52.86 & N/A & N/A & N/A & 2.04 & 1.03 & 70.59 & 10.26 & 6.26 & 28.57 & 1.36 & 0.69 & 43.10 \\
        \hline
        MNLI & \SDM & 21.24 & 13.56 & 48.98 & N/A & N/A & N/A & N/A & N/A & N/A & 23.56 & 14.46 & 63.64 & 20.54 & 12.24 & 63.64 & 31.36 & 20.92 & 62.64 & 12.99 & 8.33 & 29.41 & 18.24 & 11.36 & 46.15 \\
        \hdashline
        MNLI & DOMINO & 23.76 & 21.43 & 26.67 & N/A & N/A & N/A & N/A & N/A & N/A & 26.34 & 17.72 & 51.28 & 21.63 & 12.74 & 71.43 & N/A & N/A & N/A & 17.78 & 11.11 & 40.00 & 12.12 & 7.14 & 40.00\\
        \hdashline
        MNLI & \SDM-$Z$ &N/A & N/A & N/A & N/A & N/A & N/A & N/A & N/A & N/A & 17.60 & 11.11 & 42.31 & 15.89 & 9.23 & 57.14 & 39.58 & 41.38 & 37.93 & N/A & N/A & N/A & 14.04 & 8.33 & 44.44 \\
        \hdashline
        MNLI & \SDM-$E$ &N/A & N/A & N/A & N/A & N/A & N/A & N/A & N/A & N/A & 26.89 & 17.82 & 54.82 & 10.87 & 6.00 & 58.82 & 25.28 & 16.09 & 58.96 & N/A & N/A & N/A & 12.54 & 7.06 & 56.10 \\
        \hline
        SST-5 & \SDM & 47.73 & 42.42 & 54.55 & 5.41 & 2.78 & 100.00 & N/A & N/A & N/A & 27.87 & 17.84 & 63.64 & 18.06 & 10.40 & 68.42 & 21.91 & 13.08 & 67.50 & 24.14 & 15.08 & 60.50 & 16.46 & 9.29 & 72.31 \\
        \hdashline
        SST-5 & DOMINO & 43.66 & 38.42 & 50.55 & N/A & N/A & N/A & N/A & N/A & N/A & 29.69 & 19.37 & 63.59 & 12.63 & 6.99 & 65.91 & 25.34 & 15.71 & 65.38 & 22.31 & 13.40 & 66.48 & 13.29 & 7.33 & 70.97 \\
        \hdashline
        SST-5 & \SDM-$Z$ &36.78 & 29.09 & 50.00 & N/A & N/A & N/A & N/A & N/A & N/A & N/A & N/A & N/A & N/A & N/A & N/A & N/A & N/A & N/A & 23.53 & 100.00 & 13.33 & 28.49 & 17.65 & 73.91\\
        \hdashline
        SST-5 & \SDM-$E$ &31.00 & 31.58 & 30.43 & 12.50 & 6.67 & 100.00 & N/A & N/A & N/A & 15.38 & 18.18 & 13.33 & N/A & N/A & N/A & 37.21 & 25.81 & 66.67 & 28.39 & 17.76 & 70.73 & N/A & N/A & N/A \\
        \bottomrule
    \end{tabular}
    }
    \vspace{-5pt}
    \caption{Feature detection performance on all linguistics features}
    \label{tab:example_features}
    \vspace{-15pt}
\end{table*}
 
\begin{table*}[t!]
    \centering
    \resizebox{18cm}{!}{
    \begin{tabular}{c|c|ccc|ccc|ccc|ccc|ccc|ccc}
    \toprule
        \textbf{dataset} & \textbf{model} & \multicolumn{3}{c|}{\textbf{female}} & \multicolumn{3}{c|}{\textbf{male}} & \multicolumn{3}{c|}{\textbf{}} & \multicolumn{3}{c|}{\textbf{}} & \multicolumn{3}{c|}{\textbf{}} & \multicolumn{3}{c|}{\textbf{}}\\
        \hline
        \textbf{Metric} & & \textbf{AV} & \textbf{AH} & \textbf{AC} & \textbf{AV} & \textbf{AH} & \textbf{AC} & \textbf{AV} & \textbf{AH} & \textbf{AC} & \textbf{AV} & \textbf{AH} & \textbf{AC} & \textbf{AV} & \textbf{AH} & \textbf{AC} & \textbf{AN} & \textbf{AH} & \textbf{AC}\\
        \hline
        Jigsaw-gender & \SDM & 36.25 & 26.12 & 59.20 & 36.03 & 27.78 & 51.28 \\
        \hdashline
        Jigsaw-gender & DOMINO & 36.33 & 27.21 & 54.65 & 35.37 & 26.25 & 54.18 \\
        \hdashline
        Jigsaw-gender & \SDM-$Z$ &35.71 & 28.09 & 49.02 & 34.26 & 27.05 & 46.72 \\
        \hdashline
        Jigsaw-gender & \SDM-$E$ &43.31 & 38.89 & 48.88 & 41.29 & 23.74 & 55.89 \\
        \hline
        \textbf{} & \textbf{} & \multicolumn{3}{c|}{\textbf{Asian}} & \multicolumn{3}{c|}{\textbf{Black}} & \multicolumn{3}{c|}{\textbf{White}} & \multicolumn{3}{c|}{\textbf{Latino}} & \multicolumn{3}{c|}{\textbf{}} & \multicolumn{3}{c|}{\textbf{}}\\
        \hline
        Jigsaw-racial & \SDM & N/A & N/A & N/A & 27.65 & 20.20 & 43.83 & 36.91 & 29.95 & 48.07 & 1.63 & 0.85 & 17.65 \\
        \hdashline
        Jigsaw-racial & DOMINO & 7.25 & 3.92 & 47.76 & 29.19 & 19.54 & 57.69 & 38.89 & 29.09 & 58.64 & 50.08 & 35.56 & 88.89 \\
        \hdashline
        Jigsaw-racial & \SDM-$Z$ &15.46 & 8.60 & 76.52 & 30.41 & 22.53 & 46.74 & 38.55 & 31.44 & 49.84 & 23.92 & 16.11 & 46.43 \\
        \hdashline
        Jigsaw-racial & \SDM-$E$ &N/A & N/A & N/A & N/A & N/A & N/A & 39.12 & 33.70 & 46.61 & 35.86 & 28.82 & 47.43 \\
        \hline
        \textbf{} & \textbf{} & \multicolumn{3}{c|}{\textbf{atheist}} & \multicolumn{3}{c|}{\textbf{Buddhist}} & \multicolumn{3}{c|}{\textbf{Christian}} & \multicolumn{3}{c|}{\textbf{Hindu}} & \multicolumn{3}{c|}{\textbf{Jewish}} & \multicolumn{3}{c|}{\textbf{Muslim}}\\
        \hline
        Jigsaw-religion & \SDM & 30.00 & 27.27 & 33.33 & 80.00 & 66.67 & 100.00 & 32.76 & 24.89 & 47.89 & 61.54 & 80.00 & 50.00 & 13.82 & 9.38 & 26.32 & 34.34 & 26.85 & 47.64 \\
        \hdashline
        Jigsaw-religion & DOMINO & N/A & N/A & N/A & N/A & N/A & N/A & 33.48 & 23.96 & 55.52 & N/A & N/A & N/A & 36.29 & 28.48 & 50.00 & 34.05 & 27.72 & 44.13 \\
        \hdashline
        Jigsaw-religion & \SDM-$Z$ &N/A & N/A & N/A & 42.67 & 27.59 & 94.12 & 30.95 & 23.07 & 46.99 & 38.36 & 24.14 & 93.33 & 41.99 & 35.29 & 51.82 & 38.76 & 31.39 & 50.67 \\
        \hdashline
        Jigsaw-religion & \SDM-$E$ &N/A & N/A & N/A & 14.74 & 28.00 & 10.00 & 35.73 & 28.28 & 48.51 & 1.47 & 0.78 & 11.76 & 33.61 & 32.28 & 35.06 & 32.53 & 28.10 & 38.60 \\
        \bottomrule
    \end{tabular}
    }
    \vspace{-5pt}
    \caption{Feature detection performance on pragmatic features}
    \label{tab:significance_test}
    \vspace{-15pt}
\end{table*}

\begin{table*}[t!]
    \centering
    \resizebox{18cm}{!}{
    \begin{tabular}{c|c|ccc|ccc|ccc|ccc|ccc|ccc|ccc|ccc|ccc|c}
    \toprule
        \textbf{dataset} & \textbf{model} & \multicolumn{3}{c|}{\textbf{length}} & \multicolumn{3}{c|}{\textbf{negation}} & \multicolumn{3}{c|}{\textbf{reflexive}} & \multicolumn{3}{c|}{\textbf{CP}} & \multicolumn{3}{c|}{\textbf{NS}} & \multicolumn{3}{c|}{\textbf{MP}} & \multicolumn{3}{c|}{\textbf{quantifier}} & \multicolumn{3}{c|}{\textbf{TP}} & \multicolumn{3}{c|}{\textbf{LD}} & \multicolumn{1}{c}{}\\
        \hline
        \textbf{Metric} & & \textbf{F1} & \textbf{precision} & \textbf{recall} & \textbf{F1} & \textbf{precision} & \textbf{recall} & \textbf{F1} & \textbf{precision} & \textbf{recall} & \textbf{F1} & \textbf{precision} & \textbf{recall} & \textbf{F1} & \textbf{precision} & \textbf{recall} & \textbf{F1} & \textbf{precision} & \textbf{recall} & \textbf{F1} & \textbf{precision} & \textbf{recall} & \textbf{F1} & \textbf{precision} & \textbf{recall} & \textbf{F1} & \textbf{precision} & \textbf{recall} & \textbf{average F1}\\
        \hline
        CoLA & \SDM & 44.75 & 28.82 & \textbf{100.00} & \textbf{28.90} & \textbf{17.50} & \textbf{82.89} & \textbf{19.13} & 10.58 & \textbf{100.00} & \textbf{18.56} & \textbf{10.39} & \textbf{86.96} & 36.45 & 22.28 & \textbf{100.00} & \textbf{63.16} & 46.15 & \textbf{100.00} & 29.69 & 17.46 & \textbf{98.96} & \textbf{54.02} & \textbf{37.00} & \textbf{100.00} & \textbf{55.65} & \textbf{38.55} & \textbf{100.00} & \textbf{38.91} \\
        \hdashline
        CoLA & DOMINO & \textbf{47.45} & \textbf{32.01} & 91.67 & 26.63 & 15.97 & 80.26 & 18.86 & \textbf{11.22} & 86.27 & 16.08 & 8.94 & 80.43 & \textbf{40.24} & \textbf{26.68} & 81.82 & 60.14 & \textbf{48.01} & 80.48 & \textbf{33.98} & \textbf{20.91} & 90.63 & 44.44 & 32.85 & 68.70 & 51.38 & 37.57 & 81.25 & 37.68\\
        \hline
        QNLI & \SDM & \textbf{16.31} & \textbf{11.93} & \textbf{25.76} & 3.29 & 1.83 & 15.59 & \textbf{6.70} & \textbf{3.84} & \textbf{26.32} & \textbf{3.72} & \textbf{2.04} & \textbf{21.74} & \textbf{12.90} & \textbf{8.11} & \textbf{31.40} & \textbf{19.09} & \textbf{16.38} & \textbf{22.86} & \textbf{14.69} & \textbf{10.24} & \textbf{25.95} & 7.89 & 4.68 & \textbf{25.00} & \textbf{14.64} & 11.00 & \textbf{21.88} & \textbf{10.36} \\
        \hdashline
        QNLI & DOMINO & 13.76 & 10.57 & 19.70 & \textbf{6.00} & \textbf{3.70} & \textbf{15.79} & 4.97 & 2.89 & 17.65 & 3.29 & 1.85 & 15.22 & 10.40 & 7.42 & 17.36 & 15.74 & 15.31 & 16.19 & 11.48 & 8.94 & 16.03 & \textbf{9.27} & \textbf{5.88} & 21.88 & 14.58 & \textbf{12.50} & 17.50 & 9.94 \\
        \hline
        QQP & \SDM & \textbf{13.51} & \textbf{9.07} & \textbf{26.52} & \textbf{7.28} & \textbf{4.26} & \textbf{25.00} & \textbf{4.78} & \textbf{2.57} & \textbf{33.34} & \textbf{6.66} & \textbf{3.60} & \textbf{43.48} & \textbf{9.62} & 5.84 & \textbf{27.27} & \textbf{20.20} & 17.54 & \textbf{23.81} & 12.42 & 8.63 & \textbf{22.14} & \textbf{9.12} & \textbf{5.34} & \textbf{31.25} & \textbf{17.02} & 12.32 & \textbf{27.50} & \textbf{11.18} \\
        \hdashline
        QQP & DOMINO & 11.40 & 8.30 & 18.18 & 7.05 & 4.13 & 23.68 & 3.73 & 2.08 & 17.65 & 5.49 & 3.04 & 28.26 & 9.33 & \textbf{6.02} & 20.66 & 18.79 & \textbf{17.72} & 20.00 & \textbf{15.30} & \textbf{11.69} & \textbf{22.14} & 7.97 & 4.87 & 21.88 & 16.89 & \textbf{13.31} & 23.13 & 10.66 \\
        \hline
        SST-2 & \SDM & 10.19 & 8.79 & \textbf{12.12} & 9.49 & 6.25 & \textbf{19.74} & \textbf{2.95} & \textbf{1.82} & \textbf{7.84} & \textbf{4.10} & \textbf{2.52} & \textbf{10.87} & 10.33 & 9.33 & 11.57 & \textbf{10.79} & \textbf{16.19} & \textbf{8.10} & 8.46 & 8.53 & 8.40 & 4.89 & 3.31 & \textbf{9.38} & \textbf{19.11} & \textbf{21.05} & \textbf{17.50} & 8.92 \\
        \hdashline
        SST-2 & DOMINO & \textbf{10.73} & \textbf{10.85} & 10.61 & \textbf{10.00} & \textbf{7.07} & 17.11 & 2.70 & 1.75 & 5.88 & 3.58 & 2.15 & \textbf{10.87} & \textbf{11.80} & \textbf{9.45} & \textbf{15.70} & 9.32 & 13.39 & 7.14 & \textbf{9.77} & \textbf{9.63} & \textbf{9.92} & \textbf{5.11} & \textbf{3.69} & 8.33 & 17.45 & 20.87 & 15.00 & \textbf{8.94}\\
        \hline
        MNLI & \SDM & \textbf{15.46} & 9.71 & \textbf{37.88} & 5.87 & 3.15 & \textbf{43.42} & 4.52 & 2.46 & \textbf{27.45} & 3.42 & 1.79 & \textbf{39.13} & \textbf{12.70} & 7.69 & \textbf{36.36} & \textbf{20.63} & \textbf{15.03} & \textbf{32.86} & \textbf{7.51} & 4.16 & \textbf{38.54} & \textbf{16.77} & \textbf{11.35} & \textbf{32.06} & \textbf{15.45} & \textbf{11.76} & \textbf{22.50} & \textbf{11.37}\\
        \hdashline
        MNLI & DOMINO & 14.41 & \textbf{11.63} & 18.94 & \textbf{9.00} & \textbf{6.10} & 17.11 & \textbf{6.57} & \textbf{4.32} & 13.73 & \textbf{6.26} & \textbf{3.61} & 17.36 & 12.35 & \textbf{9.59} & 17.36 & 15.70 & 13.87 & 18.10 & 6.33 & \textbf{5.60} & 7.29 & 12.47 & 9.45 & 18.32 & 10.27 & 9.05 & 11.88 & 10.37\\
        \hline
        SST-5 & \SDM & \textbf{16.16} & \textbf{9.48} & \textbf{54.55} & \textbf{6.32} & \textbf{3.38} & \textbf{50.00} & \textbf{5.66} & \textbf{2.95} & \textbf{70.59} & \textbf{4.10} & \textbf{2.14} & \textbf{52.17} & \textbf{26.09} & \textbf{16.90} & \textbf{57.14} & \textbf{10.97} & \textbf{6.09} & 55.37 & \textbf{8.90} & \textbf{4.82} & \textbf{58.33} & 14.90 & 8.79 & \textbf{48.85} & \textbf{18.56} & 11.24 & \textbf{53.13} & \textbf{12.41} \\
        \hdashline
        SST-5 & DOMINO & 15.19 & 9.09 & 46.21 & 6.08 & 3.27 & 43.42 & 5.66 & 2.98 & 56.86 & 3.99 & 2.08 & 52.17 & 24.97 & 16.13 & 55.24 & 10.86 & 5.99 & \textbf{58.68} & 8.49 & 4.64 & 50.00 & \textbf{16.30} & \textbf{9.81} & 48.09 & 18.11 & \textbf{11.30} & 45.63 & 12.08 \\
        \bottomrule
    \end{tabular}
    }
    \vspace{-5pt}
    \caption{Synthetic feature detection precision, recall, F1 result}
    \label{tab:synthetic_exp}
    \vspace{-15pt}
\end{table*}

\begin{table*}[t!]
    \centering
    \resizebox{18cm}{!}{
    \begin{tabular}{c|c|ccc|ccc|ccc|ccc|ccc|ccc|c}
    \toprule
        \textbf{dataset} & \textbf{model} & \multicolumn{3}{c|}{\textbf{female}} & \multicolumn{3}{c|}{\textbf{male}} & \multicolumn{3}{c|}{\textbf{}} & \multicolumn{3}{c|}{\textbf{}} & \multicolumn{3}{c|}{\textbf{}} & \multicolumn{3}{c|}{\textbf{}} & \multicolumn{1}{c}{}\\
        \hline
        \textbf{Metric} & & \textbf{F1} & \textbf{precision} & \textbf{recall} & \textbf{F1} & \textbf{precision} & \textbf{recall} & \textbf{F1} & \textbf{precision} & \textbf{recall} & \textbf{F1} & \textbf{precision} & \textbf{recall} & \textbf{F1} & \textbf{precision} & \textbf{recall} & \textbf{F1} & \textbf{precision} & \textbf{recall} & \textbf{average F1}\\
        \hline
        Jigsaw-gender & \SDM & \textbf{69.67} & \textbf{56.15} & 91.76 & \textbf{68.15} & \textbf{51.69} & \textbf{100.00} & & & & & & & & & & & & & \textbf{68.91} \\
        \hdashline
        Jigsaw-gender & DOMINO & 69.38 & 53.34 & 99.04 & 67.01 & 50.58 & 99.26 & & & & & & & & & & & & & 68.20 \\
        \hline
        \textbf{} & \textbf{} & \multicolumn{3}{c|}{\textbf{Asian}} & \multicolumn{3}{c|}{\textbf{Black}} & \multicolumn{3}{c|}{\textbf{White}} & \multicolumn{3}{c|}{\textbf{Latino}} & \multicolumn{3}{c|}{\textbf{}} & \multicolumn{3}{c|}{\textbf{}} & \multicolumn{1}{c}{}\\
        \hline
        Jigsaw-racial & \SDM & \textbf{19.92} & \textbf{11.29} & 84.96 & \textbf{62.59} & \textbf{45.55} & \textbf{100.00} & 66.65 & 50.63 & 97.51 & \textbf{18.86} & \textbf{10.70} & 75.98 & & & & & & & \textbf{42.00} \\
        \hdashline
        Jigsaw-racial & DOMINO & 19.61 & 10.96 & \textbf{93.09} & 60.64 & 43.76 & 98.68 & \textbf{69.42} & \textbf{53.67} & \textbf{98.27} & 16.18 & 8.81 & \textbf{99.51} & & & & & & & 41.46\\
        \hline
        \textbf{} & \textbf{} & \multicolumn{3}{c|}{\textbf{Atheist}} & \multicolumn{3}{c|}{\textbf{Buddhist}} & \multicolumn{3}{c|}{\textbf{Christian}} & \multicolumn{3}{c|}{\textbf{Hindu}} & \multicolumn{3}{c|}{\textbf{Jewish}} & \multicolumn{3}{c|}{\textbf{Muslim}} & \multicolumn{1}{c}{}\\
        \hline
        Jigsaw-religion & \SDM & \textbf{14.44} & \textbf{7.82} & 93.70 & \textbf{11.02} & \textbf{5.85} & 94.29 & \textbf{71.15} & 56.61 & \textbf{94.08} & 9.59 & 5.07 & 87.50 & \textbf{54.74} & \textbf{38.89} & 92.42 & 59.29 & 42.63 & \textbf{97.33} & \textbf{36.71} \\
        \hdashline
        Jigsaw-religion & DOMINO & 11.59 & 6.15 & \textbf{100.00} & 9.54 & 5.01 & \textbf{97.14} & 70.19 & \textbf{58.77} & 87.10 & \textbf{9.77} & \textbf{5.24} & 71.88 & 54.13 & 37.56 & \textbf{96.88} & \textbf{59.45} & \textbf{44.36} & 90.13 & 35.77\\
        \bottomrule
    \end{tabular}
    }
    \vspace{-5pt}
    \caption{Synthetic feature detection F1 result on Jigsaw}
    \label{tab:synthetic_exp_jigsaw}
    \vspace{-15pt}
\end{table*}

\begin{table*}[!ht]
    \centering
    \resizebox{18cm}{!}{
    \begin{tabular}{c|ccc|ccc|ccc|ccc|ccc|ccc|ccc|ccc|ccc|c}
    \toprule
        \textbf{} & \multicolumn{3}{c|}{\textbf{length}} & \multicolumn{3}{c|}{\textbf{negation}} & \multicolumn{3}{c|}{\textbf{reflexive}} & \multicolumn{3}{c|}{\textbf{CP}} & \multicolumn{3}{c|}{\textbf{NS}} & \multicolumn{3}{c|}{\textbf{MP}} & \multicolumn{3}{c|}{\textbf{quantifier}} & \multicolumn{3}{c|}{\textbf{TP}} & \multicolumn{3}{c|}{\textbf{LD}} & \multicolumn{1}{c}{}\\
        \hline
        \textbf{Metric} & \textbf{F1} & \textbf{precision} & \textbf{recall} & \textbf{F1} & \textbf{precision} & \textbf{recall} & \textbf{F1} & \textbf{precision} & \textbf{recall} & \textbf{F1} & \textbf{precision} & \textbf{recall} & \textbf{F1} & \textbf{precision} & \textbf{recall} & \textbf{F1} & \textbf{precision} & \textbf{recall} & \textbf{F1} & \textbf{precision} & \textbf{recall} & \textbf{F1} & \textbf{precision} & \textbf{recall} & \textbf{F1} & \textbf{precision} & \textbf{recall} & \textbf{average F1}\\
        \hline
         & 44.75 & 28.82 & \textbf{100.00} & \textbf{28.90} & \textbf{17.50} & 82.89 & \textbf{19.13} & \textbf{10.58} & \textbf{100.00} & 18.56 & 10.39 & \textbf{86.96} & \textbf{36.45} & 22.28 & \textbf{100.00} & 63.16 & 46.15 & \textbf{100.00} & 29.69 & 17.46 & 98.96 & 54.02 & 37.00 & \textbf{100.00} & 55.65 & 38.55 & \textbf{100.00} & \textbf{38.91}\\
        \hline
        \hline
        $\gamma = 0.5$ & 48.26 & 37.34 & 68.18 & 21.50 & 12.35 & 82.89 & 15.94 & 8.74 & 90.20 & \textbf{23.81} & \textbf{14.11} & 76.9 & 30.63 & 18.80 & 82.64 & 52.34 & 36.02 & 95.71 & 26.31 & 15.51 & 86.46 & 40.51 & 28.11 & 72.52 & 52.28 & 35.80 & 96.88 & 34.62 \\
        \hline
        $\gamma = 1$ & \textbf{64.71} & \textbf{56.90} & 75.00 & 24.74 & 14.71 & 77.63 & 10.00 & 5.66 & 43.14 & N/A & N/A & N/A & 29.44 & \textbf{38.16} & 23.97 & 56.05 & 50.57 & 62.86 & 25.87 & 15.67 & 73.95 & \textbf{66.43} & \textbf{61.84} & \textbf{71.76} & 37.23 & 25.11 & 71.88 & 34.94\\
        \hline
        \hline
        $\gamma_C = 0$ & 46.79 & 30.61 & 99.24 & 28.57 & 17.26 & 82.89 & 17.68 & 9.70 & \textbf{100.00} & 18.69 & 10.47 & \textbf{86.96} & 36.39 & 22.24 & \textbf{100.00} & \textbf{63.25} & \textbf{46.26} & \textbf{100.00} & \textbf{30.26} & \textbf{17.97} & 95.83 & 52.94 & 36.00 & \textbf{100.00} & 54.12 & 37.94 & 94.38 & 35.08\\
        \hline
        $\gamma_C = 0.5$ & 47.40 & 31.06 & \textbf{100.00} & 27.05 & 15.63 & \textbf{100.00} & 17.92 & 9.86 & 98.04 & 18.48 & 10.34 & \textbf{86.96} & 34.97 & 21.19 & \textbf{100.00} & 60.87 & 43.75 & \textbf{100.00} & 29.54 & 17.33 & \textbf{100.00} & 53.69 & 36.69 & \textbf{100.00} & \textbf{57.30} & \textbf{40.46} & 98.13 & 38.58 \\
        \hline
        $\gamma_C = 1$  & 47.40 & 31.06 & \textbf{100.00} & 26.76 & 15.45 & \textbf{100.00} & 15.69 & 8.51 & \textbf{100.00} & 20.89 & 11.87 & \textbf{86.96} & N/A & N/A & N/A & 56.78 & 42.45 & 85.71 & 29.31 & 17.17 & \textbf{100.00} & 52.19 & 35.31 & \textbf{100.00} & 54.70 & 37.65 & \textbf{100.00} & 33.75\\
        \bottomrule
    \end{tabular}
    }
    \vspace{-5pt}
    \caption{Ablation study on synthetic feature detection on CoLA dataset}
    \label{tab:synthetic_ablation}
    \vspace{-15pt}
\end{table*}
\fi

\end{document}